\documentclass{article}
\pdfoutput=1

\usepackage{microtype}
\usepackage{graphicx}
\usepackage{subcaption}
\usepackage{booktabs} 

\usepackage{hyperref}

\usepackage[accepted]{icml2026}

\usepackage{amsmath}
\usepackage{amssymb}
\usepackage{mathtools}
\usepackage{amsthm}
\usepackage{bm}
\usepackage{multirow}

\usepackage[capitalize,noabbrev]{cleveref}

\theoremstyle{plain}

\theoremstyle{definition}

\theoremstyle{remark}

\usepackage{algorithm}
\usepackage{algorithmic}

\DeclareMathOperator{\Cov}{Cov}
\DeclareMathOperator{\softmax}{softmax}

\newcommand{\ie}{i.e.,\ }
\newcommand{\eg}{e.g.,\ }

\newcommand{\mubar}{\overline{\mu}}

\newcommand{\qbar}{\overline{q}}
\newcommand{\qres}{\tilde{q}}
\newcommand{\kbar}{\overline{k}}
\newcommand{\kres}{\tilde{k}}
\newcommand{\vbar}{\overline{v}}

\newcommand{\DQ}{C_q}
\newcommand{\DK}{C_k}
\newcommand{\DC}{C}

\newcommand{\DS}{N}

\newcommand{\DD}{D}

\newcommand{\ours}{MuSe}
\newcommand{\Skew}{\operatorname{Skew}}

\icmltitlerunning{Multipole Semantic Attention}

\begin{document}

\twocolumn[
\icmltitle{Multipole Semantic Attention: A Fast Approximation of \\ Softmax Attention for Pretraining}

\icmlsetsymbol{equal}{*}

\begin{icmlauthorlist}
\icmlauthor{Rupert Mitchell}{tu}
\icmlauthor{Kristian Kersting}{tu,hessian,dfki,cog}
\end{icmlauthorlist}

\icmlaffiliation{tu}{Department of Computer Science, TU Darmstadt, Darmstadt, Germany}
\icmlaffiliation{hessian}{Hessian Center for AI (hessian.AI), Darmstadt, Germany}
\icmlaffiliation{dfki}{German Research Center for Artificial Intelligence (DFKI), Darmstadt, Germany}
\icmlaffiliation{cog}{Center for Cognitive Science, TU Darmstadt, Darmstadt, Germany}

\icmlcorrespondingauthor{Rupert Mitchell}{mail@rupertmitchell.com \mbox{(institutional: rupert.mitchell@tu-darmstadt.de)}}

\icmlkeywords{Machine Learning, ICML, transformers, attention mechanisms, efficient attention, long context, pretraining}

\vskip 0.3in
]

\printAffiliationsAndNotice{}

\begin{abstract}
Pretraining transformers on long sequences, such as entire code repositories or collections of related documents, is bottlenecked by quadratic attention costs.
We present Multipole Semantic Attention (\ours{}), which accelerates 64k-context pretraining by 36\% while matching baseline loss, requiring no architectural changes.
\ours{} is a training-time approximation that clusters queries and keys separately in representation space.
This yields query-specific summaries that substantially outperform spatial blocking at matched sparsity, while also enabling drop-in compatibility with existing pretrained models---we validate on Llama 3.1-8B and 3.2-1B without retraining.
We pretrain language models up to 1B parameters at 64k context on code and scientific documents, confirming that \ours{} preserves quality and long-context utilization during training.
\end{abstract}

\section{Introduction}
\label{sec:introduction}

The quadratic computational complexity of softmax attention remains the primary bottleneck limiting context length in transformers. While this $\mathcal{O}(\DS^2\DD)$ scaling enables the rich token interactions that underpin transformer capabilities, it renders long-context pretraining prohibitively expensive. Modern pretraining increasingly benefits from extended context---whether processing entire code repositories or collections of related scientific papers---yet computational constraints force most models to train on artificially truncated sequences. Modern transformers partially mitigate these issues through the use of flash attention~\citep{dao2022_flashattention}, which reduces memory complexity to linear but retains quadratic computational complexity. Hybrid architectures using sliding window attention for local interactions must still interleave quadratic-complexity global attention layers to maintain full-context understanding.
Efficient approximations to global attention, which maintain training quality at these scales, become essential for practical long-context pretraining.

In this context, we present Multipole Semantic Attention (\ours), combining semantic clustering with ideas from computational physics to approximate softmax attention at training time. We demonstrate that (1) the approximation matches or exceeds baseline training quality, (2) throughput improvements of 36\% are achievable with current hardware at context lengths of 64k, and (3) the method integrates into existing training pipelines without architectural changes. More specifically, \ours{} introduces three key elements: First, we cluster queries and keys separately in their learned representation spaces, enabling a two-stage mechanism where coarse query clusters attend to fine key clusters, then fine queries refine through cluster summaries. Second, we augment centroid-based (monopole) approximations with retrieval of the most relevant clusters for exact attention. Third, query centroids provide exponential tilts that center the approximation around each query cluster's region of attention, enabling drop-in compatibility with existing pretrained models without retraining.

This distinction between the key and query latent spaces, whereby we cluster keys and queries independently, is motivated as follows.
Firstly, softmax attention is invariant under translation of the keys (as this change is absorbed by the normalizing constant), but not of the queries.
More importantly, the queries live in the dual vector space to the keys, that is, softmax attention is further invariant to an arbitrary change of basis of the keys, so long as the queries are transformed inversely.
The consequence of this is that, unless one has some way to choose a preferred basis, it is unclear what it would mean to say that some particular key and query occupied the same point in latent space.

We validate \ours{} empirically through microbenchmarks and end-to-end pretraining at scale. On isolated attention layers, we achieve 2$\times$ speedup compared to CUDNN Flash Attention at 64k context with approximation errors below 1\%. We pretrain 1B parameter models on 64k context using both code repositories and scientific documents, demonstrating 36\% wall-clock speedup while matching baseline quality. We further validate that the method generalizes to existing pretrained models (Llama 3.1-8B and 3.2-1B) and confirm that our models utilize the full 64k context during training.

To summarize, this paper makes the following contributions:
\begin{itemize}
\item A bidirectional semantic clustering approach that partitions queries and keys \textit{separately} in their native representation space, enabling a hierarchical two-stage attention mechanism: coarse query clusters first attend to all fine key clusters to produce query-dependent summaries, then fine queries refine these summaries using their residual components
\item A retrieval-augmented approximation that selects the most relevant key-value clusters for exact attention, with causal accumulation of summaries across spatial blocks
\item Empirical validation at 1B parameter scale on code and scientific documents with 64k context, achieving 36\% wallclock speedup while maintaining training quality
\end{itemize}

\section{Related Work}
\label{sec:related_work}

We organize existing efficient attention methods into two broad categories: those which restrict attention to subsets of tokens exactly, and those which approximate attention globally.
We fall into the latter category, and together with \citet{hooper2025_multipoleattention} are the only works using semantic clustering therein.

\begin{figure}[!t]
\centering
\includegraphics[width=\columnwidth]{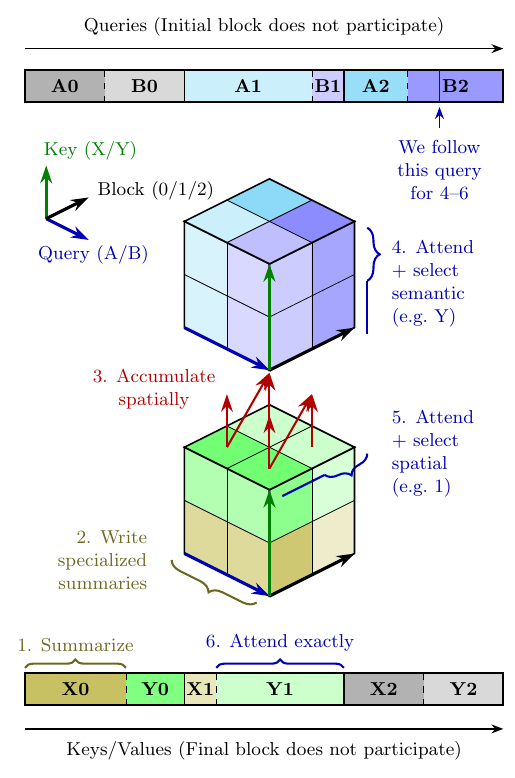}
\caption{\ours{} method overview, depicting the far-field approximation.
\textbf{Bars:} Queries (top) and keys/values (bottom) are partitioned by semantic cluster (A/B for queries, X/Y for keys) within each spatial block (0, 1, 2); segments vary in size because cluster membership is data-dependent.
\textbf{Lower cube:} Per-block summaries indexed by (query cluster, key cluster, spatial block), constructed in Steps 1--2.
\textbf{Upper cube:} Causally accumulated summaries (Step 3), accumulated along the spatial axis.
\textbf{Steps 4--6:} A query in segment B2 (blue line) attends to the upper cube to select key cluster Y (Step 4, semantic selection), then to the lower cube to select spatial block 1 (Step 5, spatial selection), and finally attends exactly to keys in (Y, 1) (Step 6).
Greyed-out segments do not participate: the first query block has no prior context, and the final key/value block is only accessed via local attention (computed separately).
See Algorithm~\ref{alg:muse} for pseudocode.}
\label{fig:method-overview}
\end{figure}

\paragraph{Restricted Attention}
Reformers~\citep{kitaev2020_reformer} bucket queries and keys according to random directions and restrict attention to within buckets.
Routing Transformers~\citep{roy2021_routingtransformer} learn a k-means clustering during training and similarly restrict attention to within buckets.
Since late 2024 there has been significant work on K-means clustering in key-space for inference acceleration: \citet{hooper2024_squeezedattention} use context-specific clustering to perform exact attention on the most relevant keys, while Tactic~\citep{zhu2025_tactic} adapts the number of retrieved tokens to attention temperature.

\paragraph{Approximate Attention}
\citet{wang2020_linformer} use random projections to approximate the attention matrix, exploiting its low rank structure.
Nystr\"omformer~\citep{xiong2021_nystromformer} constructs a low rank representation using segment-wise means as landmarks.
Fast Multipole Attention~\citep{fastmultipole2023} and H-Transformer~\citep{zhu2021_htransformer} use recursive spatial decomposition inspired by Barnes-Hut simulation~\citep{barneshut1986}.
Performer~\citep{Choromanski2021_performer} uses random features to approximate the exponential kernel.
Mixture of Blocks Attention \citep[MoBA;][]{lu2025moba} uses fixed spatial blocks without semantic clustering.

\paragraph{Positioning of MuSe}
Our work uses semantic clustering of key-query representation space.
Unlike sparse training methods such as MoBA that attend to fixed spatial blocks, we compute summaries of all clusters and augment with retrieval of high-attention clusters for exact computation.
Unlike \citet{hooper2025_multipoleattention} who apply monopole summaries at inference time with key clustering only, we target pretraining where massively parallel queries make query clustering practical: computing query-specific summaries amortizes well while substantially improving approximation quality.
Crucially, \ours{} requires no architectural changes: the approximation is compatible with pretrained models (validated on Llama) and \ours{}-trained models switch to exact attention with minimal adaptation.

\section{Multipole Semantic Attention (\ours{})}
\label{sec:method}

\begin{figure}[!t]
\centering
\includegraphics[width=0.62\columnwidth]{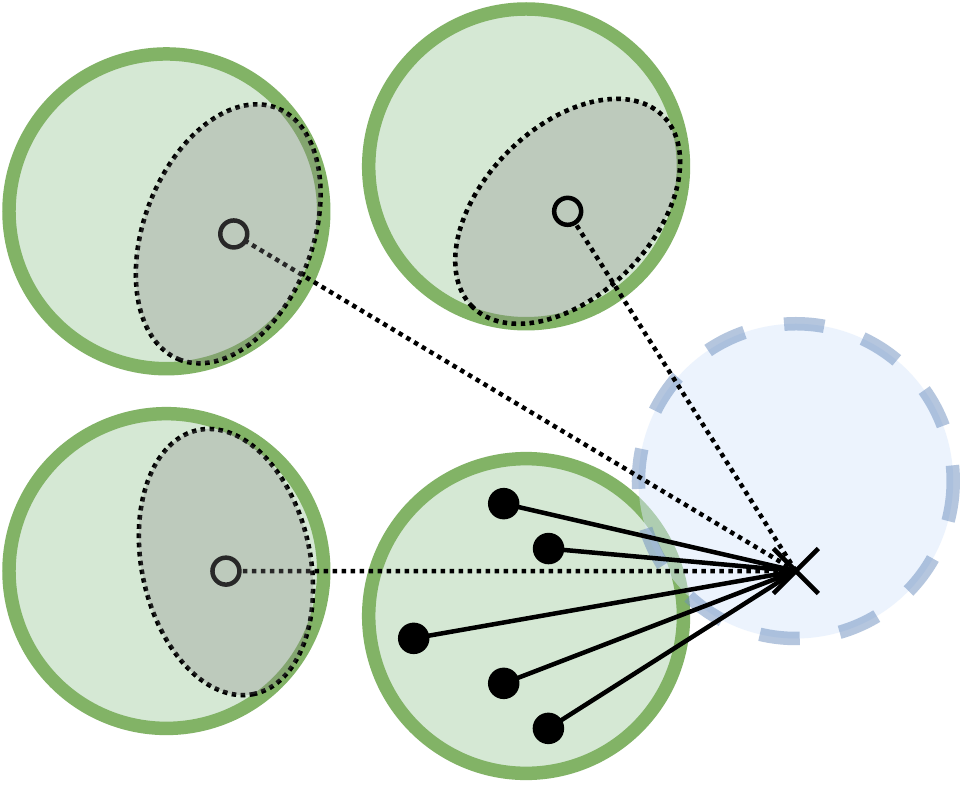}
\caption{Geometric interpretation of \ours{}. Green circles represent key/value clusters; the blue dashed circle represents a query cluster. Hollow dots mark the exponentially-tilted centroids---cluster summaries shifted toward the query cluster. Solid dots mark individual key/value pairs selected for exact retrieval. The chosen query ($\times$) connects to tilted centroids via dotted lines (approximate attention) and to retrieved tokens via solid lines (exact attention).}
\label{fig:geometric}
\end{figure}

Softmax attention is quadratic because every query attends to every key.
The core idea of \ours{} is to group both queries and keys by ``semantic'' similarity---similarity in the post-position-encoding representation space seen by attention---rather than solely by sequence position (key clusters depicted as green circles in Figure~\ref{fig:geometric}).
Clustering queries separately from keys enables query-specific summaries that tightly approximate attention while remaining interchangeable with exact attention at test time.
We combine this with spatial blocking for causality and selective retrieval of important (cluster, block) pairs for exact attention.

The core data structure is a summaries tensor of shape $\DQ \times \DK \times (N/B)$, where $\DQ$ and $\DK$ are the number of query and key clusters (typically $\DQ = \DK = \DC$), and $N/B$ is the number of spatial blocks.
Each entry summarizes the keys and values in a specific (query cluster, key cluster, spatial block) triple, enabling efficient approximate attention with selective exact retrieval.

\paragraph{Spatial blocks}
The context of length $N$ is divided into $N/B$ spatial blocks of size $B$ (\eg 8 blocks of 8k tokens for 64k context).
This structure enables causal masking at block granularity and provides the second level of retrieval.

\paragraph{Clustering}
Queries and keys are clustered \emph{globally} across the full context\footnote{Clustering and retrieval selection are recomputed from the current queries and keys on every forward pass; \ours{} maintains no state across sequences or training steps.} using a single pass of mini-batch K-means followed by a final assignment (see Appendix~\ref{app:clustering}), ignoring block boundaries.
Each query $q$ decomposes as $q = \qbar_i + \qres$ where $\qbar_i$ is the centroid of its assigned cluster $i$ and $\qres$ is the residual.
Clustering queries separately from keys---rather than clustering only keys as in prior work---is a central contribution.

Standard softmax attention computes, for each query $q$:
\begin{align}
    Z &= \textstyle\sum_k \exp(q \cdot k) \\
    o &= Z^{-1} \textstyle\sum_k \exp(q \cdot k) \, v_k
\end{align}
As we show next, \ours{} approximates the majority of these exponential weights by dropping the interaction between query and key residuals.
The result is accurate provided that any weights carrying large fractional error contribute negligibly to it.

\paragraph{Why separate query clustering helps?}
Consider the attention weight $\exp(q \cdot k)$ with $q = \qbar + \qres$ and $k = \kbar + \kres$.
Expanding the dot product: $\exp(q \cdot k) =$
\begin{equation}
\label{eq:monopole}
    \exp\bigl(
    \qbar \cdot \kbar +
    \qbar \cdot \kres +
    \qres \cdot \kbar +
    \qres \cdot \kres
    \bigr)
    \approx
    \exp\bigl(
    \qbar \cdot \kbar +
    \qbar \cdot \kres +
    \qres \cdot \kbar
    \bigr)
\end{equation}
If we cluster only keys (approximating $k \approx \kbar$), we lose the two terms involving $\kres$.
With both query and key clustering, our two-stage mechanism retains three of four terms, dropping only $\qres \cdot \kres$---the product of two residuals.
On our 1B model, this tighter approximation reduces monopole error by $3.7$--$5.6\times$ (Table~\ref{tab:decomposition}). Table~\ref{tab:1b_effective} confirms both that query clustering provides a ${\sim}9\times$ effective cluster count advantage, and that finer clustering consistently reduces error by shrinking residual magnitudes.

\subsection{Summary Computation and Causal Accumulation}
\label{sec:summaries}

We compress keys and values into compact cluster summaries---the monopole approximation---specialized for each query cluster via exponential tilting.
These are then accumulated causally across spatial blocks.

\paragraph{Per-block summaries}
For each spatial block $b$, we compute summaries for all (query cluster $i$, key cluster $j$) pairs by attending from the query centroid $\qbar_i$ to all keys in block $b$ that belong to cluster $j$.
Let $S_{ijb} = \sum_{k \in (b,j)} \exp(\qbar_i \cdot k)$ denote the unnormalized attention mass. The summaries are:
\begin{align}
    \kbar_{ijb} &= S_{ijb}^{-1} \textstyle\sum_{k \in (b,j)} \exp(\qbar_i \cdot k) \, k \\
    \vbar_{ijb} &= S_{ijb}^{-1} \textstyle\sum_{k \in (b,j)} \exp(\qbar_i \cdot k) \, v \label{eq:vbar} \\
    \mubar_{ijb} &= \log S_{ijb}
\end{align}
These are ``exponentially tilted'' centroids---weighted by attention from the query centroid, not simple averages (shown as hollow dots in Figure~\ref{fig:geometric}, shifted toward the query cluster).
This constitutes exact attention from coarse queries to fine keys.
The cost is $\mathcal{O}(\DC \cdot N \cdot D)$---linear in context length, compared to $\mathcal{O}(N^2 D)$ for exact attention---and produces $(N/B) \times \DQ \times \DK$ summaries (Steps 1--2 in Figure~\ref{fig:method-overview}).

\paragraph{Causal accumulation}
To enable causal attention, we accumulate summaries along the spatial axis so that block $b$ has access to summaries of all preceding blocks $0, \ldots, b-1$.
The accumulation is \emph{exclusive}---a block does not include itself:
\begin{itemize}
    \item $\mubar$: cumulative logsumexp
    \item $\kbar$, $\vbar$: cumulative softmax-weighted sum (weights derived from accumulated $\mubar$)
\end{itemize}
After accumulation, $\text{accumulated}_{ijb}$ summarizes all keys/values in blocks $0 \ldots b-1$ belonging to key cluster $j$, specialized for query cluster $i$ via exponential tilting (Step 3 in Figure~\ref{fig:method-overview}).

We implement this sequentially rather than with parallel scan (prefix sum), since there is already massive parallelism from $\DQ \times \DK \times \text{heads} \times \text{batch}$ independent accumulations.

\subsection{Two-Level Retrieval}
\label{sec:retrieval}

The monopole approximation in Eq.~\eqref{eq:monopole} drops the residual cross term $\qres \cdot \kres$.
This error is significant only when the residual query has large positive dot product with keys in a cluster.
Retrieval selects exactly these clusters for exact token-level attention; for the remaining clusters, keys either carry negligible attention weight or have small residual interaction, so the monopole approximation holds.
Intuitively, if the attention pattern is smooth, weighted centroids approximate it well; if it is sharp, the high-weight keys are few and retrieval captures them.
This decomposition---approximating smooth interactions, computing sharp ones exactly---mirrors the Fast Multipole Method~\citep{rokhlin1985_fmm, greengard1987_fmm}, with relevance in representation space playing the role of spatial distance.
This proceeds in two stages: first selecting \emph{clusters}, then selecting \emph{spatial blocks} within those clusters.

\paragraph{Cluster selection}
For each query $q = \qbar_i + \qres$ in spatial block $b$, we compute attention scores to all $\DK$ accumulated cluster summaries:
\begin{equation}
\label{eq:score}
    \text{score}_j = \qres \cdot \kbar_{ijb} + \mubar_{ijb}
\end{equation}
This is the best available estimate of the log attention mass that query $q$ will receive from cluster $j$.
We select the top-$k_1$ clusters for exact retrieval (\eg $k_1 = 8$ out of $\DC = 128$), mask them out pre-softmax, and compute approximate attention over the remaining $\DC - k_1$ clusters using the scores of Eq.~\eqref{eq:score} and value summaries of Eq.~\eqref{eq:vbar} (Step 4 in Figure~\ref{fig:method-overview}).

\paragraph{Spatial block selection}
For queries that selected cluster $j$, we now select which spatial blocks within cluster $j$ to retrieve.
This uses the \emph{pre-accumulation} per-block summaries rather than the accumulated ones.

Queries are segmented by which cluster(s) they are retrieving.
For queries in query cluster $i$ that selected key cluster $j$, we attend to the $N/B$ per-block summaries for the $(i, j)$ pair, apply a block-index causal mask ($b_q > b_k$), and select the top-$k_2$ spatial blocks (\eg $k_2 = 1$ out of $N/B = 8$).
Selected blocks are masked out pre-softmax, and we compute approximate attention over the remaining blocks as above, but with \emph{pre-accumulation} summaries (Step 5 in Figure~\ref{fig:method-overview}).

The output specifies $k_1 \times k_2$ (cluster, block) pairs per query for exact retrieval (\eg $8 \times 1 = 8$ pairs).

\paragraph{Dipole corrections}
An alternative to retrieval is to improve the monopole (centroid) approximation with higher-order terms.
By expanding attention as a polynomial in $\qres$, one obtains a dipole correction involving the covariance $\Cov(v, k)$ within each cluster, reducing approximation error from $\mathcal{O}(\qres \cdot \kres)$ to $\mathcal{O}(\qres^2 \cdot \kres^2)$.
We derive this in Appendix~\ref{app:dipole}.
In practice, retrieval provides larger accuracy gains at acceptable computational cost (see Appendix~\ref{app:dipole} for an empirical comparison), so we use retrieval in our main experiments.

\subsection{Exact Retrieval and Local Attention}
\label{sec:exact}

The retrieved (cluster, block) pairs and the local block diagonal are computed with exact flash attention, recovering token-level detail that the monopole summaries cannot capture.

\paragraph{Exact retrieval}
We perform exact flash attention from each query to its selected (cluster, block) pairs, using the \emph{full} query (not the residual) (Step 6 in Figure~\ref{fig:method-overview}).
Keys and values are bucketed by (cluster, block).
We build an inverse index mapping each (cluster, block) to the queries that retrieve it, then perform segmented flash attention.

To avoid $k_1 \times k_2$ memory blowup, we use index arrays into the original queries, keys, and values rather than materializing expanded arrays, similar to the segmentation approach in FlashMoBA~\citep{xiao2025flashmoba}.

\paragraph{Local attention}
For within-block attention (the ``diagonal''), we use exact flash attention with standard causal masking.
This handles local interactions where attention is typically strongest.
(Figure~\ref{fig:method-overview} depicts only the far-field approximation; local attention is computed separately.)

\paragraph{Output merging}
The final output merges four components via logsumexp weighting:
\begin{enumerate}
    \item Approximate attention from non-retrieved clusters (from cluster selection)
    \item Approximate attention from non-retrieved blocks within retrieved clusters (from block selection)
    \item Exact attention from retrieved (cluster, block) pairs
    \item Exact local attention within the current block
\end{enumerate}

\paragraph{Complexity}
At our operating point (64k context, 8k block diagonal), the block diagonal contributes 1/8 of full quadratic cost, with the far-field approximated at 64$\times$ sparsity.
Maintaining these ratios as context scales, we find approximation error decreases substantially (Appendix~\ref{app:farfield_scaling}); quantifying the resulting speedups at longer contexts is future work.
With typical parameters ($N = 2^{16}$, $\DC = 128$, $B = 8192$, $k_1 = 8$, $k_2 = 1$), we achieve significant speedups both in theoretical FLOPs and in wall-clock time (Section~\ref{sec:experiments}).
We implement \ours{} in JAX with custom Pallas kernels for summary computation and segmented retrieval, and CUDA kernels for K-means clustering (Appendix~\ref{app:clustering}).
The approximation is compatible with standard distributed training strategies (Appendix~\ref{app:distributed}).

\section{Experiments}
\label{sec:experiments}


\subsection{Setup and Microbenchmarks}
\label{sec:microbenchmarks}


\paragraph{Runtime Comparison}
Table~\ref{tab:runtime_comparison} compares attention runtime across methods on the 1B model (64 heads, 2 sequences of 64k context each).
\ours{} achieves $2\times$ speedup over CUDNN Flash Attention while maintaining high approximation quality.
We also compare against MoBA at matched far-field sparsity: both methods use an 8k block diagonal, and MoBA retrieves one 512-token block per query to match \ours{}'s retrieval of 8 clusters $\times$ 1 spatial block (${\sim}$512 tokens).
MoBA is slightly faster than \ours{} (no summarization overhead), but \ours{} achieves $40\times$ lower approximation error (0.006 vs 0.248 relative squared error).

\begin{table}[h]
\centering
\small
\caption{\textbf{MuSe achieves 2$\times$ speedup with 40$\times$ lower error than MoBA at matched sparsity.} Attention runtime comparison (1B model, 64 heads, 2$\times$64k context). $^\dagger$Matched sparsity: 8k block diagonal, 64$\times$ far-field.}
\label{tab:runtime_comparison}
\begin{tabular}{lccc}
\toprule
Method & Time (ms) & Speedup & RSE \\
\midrule
CUDNN Flash & 225.6 & 1.00$\times$ & 0 (exact) \\
Pallas Flash & 265.6 & 0.85$\times$ & 0 (exact) \\
\midrule
\ours{} (Ours)$^\dagger$ & 114.1 & 1.98$\times$ & \textbf{0.006} \\
MoBA$^\dagger$ & 109.0 & 2.07$\times$ & 0.248 \\
Block diag.\ only & 35.0 & 6.45$\times$ & 0.481 \\
\bottomrule
\end{tabular}
\end{table}

\paragraph{1B Model Validation}
Table~\ref{tab:1b_effective} validates the query clustering benefit on our headline 1B parameter model (320 heads, head dimension 64).
We vary cluster count jointly (Q=K) with retrieval fraction fixed at R/QK = 1/16.
For each configuration, we compute the effective no-query-clustering cluster count---the cluster count the ablated method would require to achieve the same error, interpolated via power-law fit.
MuSe exhibits a steeper power-law slope ($d \approx -0.89$) than the no-query-clustering ablation ($d \approx -0.59$), so the effective cluster multiplier grows with cluster count: from $5.1\times$ at QK16 to $10.7\times$ at QK512.
At our operating point of QK128R8, MuSe achieves error that would require ${\sim}1200$ clusters without query clustering---a $9.2\times$ effective cluster count advantage.

\begin{table}[h]
\centering
\small
\caption{\textbf{Query clustering provides a ${\sim}9\times$ effective cluster count advantage.} MuSe approximation quality on 1B model (320 heads, 2 sequences, 64k context). R = QK/16 throughout. Timing measured on 40 heads due to memory constraints. Effective No-Q cluster count computed via power-law interpolation; values marked with $\dagger$ are extrapolated beyond measured No-Q data (QK512).}
\label{tab:1b_effective}
\begin{tabular}{lccccc}
\toprule
QK & RSE & Corr & Eff.\ No-Q & Mult.\ & Time (ms) \\
\midrule
16   & 0.03797 & 0.9775 & 82            & 5.1$\times$  & --- \\
32   & 0.02100 & 0.9878 & 191           & 6.0$\times$  & 130.0 \\
64   & 0.01139 & 0.9935 & 568$\dagger$  & 8.9$\times$  & 88.5 \\
\textbf{128}  & \textbf{0.00622} & \textbf{0.9965} & \textbf{1179}$\dagger$ & \textbf{9.2}$\times$  & \textbf{80.5} \\
256  & 0.00330 & 0.9980 & 2518$\dagger$ & 9.8$\times$  & 106.8 \\
512  & 0.00173 & 0.9990 & 5481$\dagger$ & 10.7$\times$ & 219.8 \\
\bottomrule
\end{tabular}
\end{table}

\paragraph{Monopole vs Retrieval Decomposition}
To understand where query clustering helps, we measure approximation quality for monopole-only (no retrieval) and retrieval-only (no monopole) variants on the 1B model.
Table~\ref{tab:decomposition} shows that query clustering dramatically improves monopole quality ($3.7$--$5.6\times$ error reduction) but barely affects retrieval selection ($1.02\times$).
This confirms that query clustering improves the \emph{quality} of cluster summaries; retrieval selection works well either way because it only needs to identify high-attention clusters, not compute precise values.

\begin{table}[h]
\centering
\small
\caption{\textbf{Query clustering acts through improved cluster summaries, not retrieval selection.} Monopole vs retrieval decomposition on 1B model. ``With Q'' uses query clustering; ``No Q'' uses zeroed query centroids. Query clustering provides $4.5\times$ benefit for monopole but only $1.02\times$ for retrieval.}
\label{tab:decomposition}
\begin{tabular}{llccc}
\toprule
Component & Setting & RSE & Corr & Q benefit \\
\midrule
\multirow{2}{*}{Monopole (K64)} & With Q & 0.0342 & 0.9795 & \multirow{2}{*}{3.7$\times$} \\
& No Q & 0.1278 & 0.9282 & \\
\midrule
\multirow{2}{*}{Monopole (K128)} & With Q & 0.0227 & 0.9866 & \multirow{2}{*}{4.5$\times$} \\
& No Q & 0.1016 & 0.9426 & \\
\midrule
\multirow{2}{*}{Monopole (K256)} & With Q & 0.0148 & 0.9913 & \multirow{2}{*}{5.6$\times$} \\
& No Q & 0.0824 & 0.9534 & \\
\midrule
\multirow{2}{*}{Retrieval (K128R8)} & With Q & 0.2429 & 0.9407 & \multirow{2}{*}{1.02$\times$} \\
& No Q & 0.2475 & 0.9367 & \\
\bottomrule
\end{tabular}
\end{table}

\begin{figure*}[t]
\centering
\begin{subfigure}[b]{0.34\textwidth}
\includegraphics[width=\textwidth]{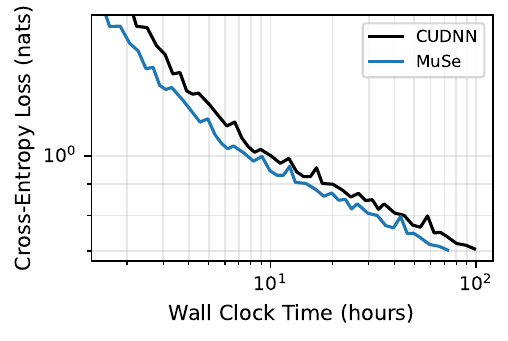}
\caption{Code}
\end{subfigure}
\hspace{0.25\columnwidth}
\begin{subfigure}[b]{0.34\textwidth}
\includegraphics[width=\textwidth]{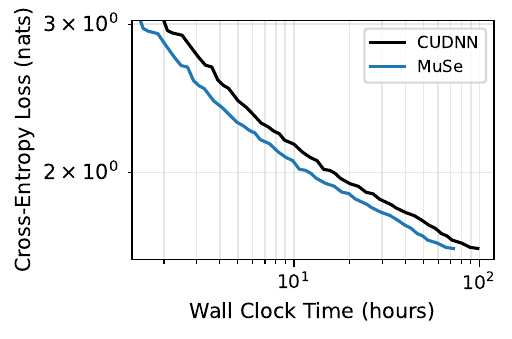}
\caption{Scientific PDFs}
\end{subfigure}
\caption{Training loss versus wall-clock time for 1B models on code (left) and scientific PDF (right) domains. \ours{} (blue) and CUDNN (black) follow the same loss trajectory, shifted horizontally---the gap is the 36\% throughput advantage. Plots start after warmup (4000 steps). Trained on a single node of 8 NVIDIA A100 GPUs.}
\label{fig:training_curves}
\end{figure*}

\subsection{Pretraining Results}
\label{sec:pretraining}

%
%
%
%

\paragraph{Headline Results}

Table~\ref{tab:headline} summarizes our main 1B parameter results on both code and scientific PDF domains.
On code, \ours{} achieves a train-MuSe/test-MuSe loss of 0.7001 compared to the CUDNN baseline of 0.7026 (0.4\% improvement).
When tested with CUDNN attention, there is a small adaptation gap (0.7108, 1.2\% degradation); however, this gap closes rapidly with brief CUDNN fine-tuning: after just 26M tokens (${\sim}0.1\%$ of pretraining), the fine-tuned model achieves 0.6994, \emph{beating} the baseline (see Appendix~\ref{app:finetuning} for details).
We further identify the gradient pathway responsible for this adaptation, namely gradients through the tilted aggregation weights of the monopole summary, and preliminarily find that making it non-differentiable during training removes the gap with no fine-tuning at all, reaching 0.6991 under exact attention (Appendix~\ref{app:clustering}, Table~\ref{tab:gradfix}).\footnote{Identified during review of this work; our main results retain the fine-tuning workflow.}
On scientific PDFs, \ours{} outperforms the baseline: 1.6166 vs 1.6201 (0.2\% improvement), with minimal interchangeability gap (1.6188 when tested with \ours{} attention).
We hypothesize two competing effects: the approximation acts as implicit regularization (improving generalization), while extended training allows minor adaptation to the approximation's specific behavior.
The regularization effect dominates early and at smaller scales, explaining why \ours{} beats the baseline; adaptation emerges with extended training but is easily removed via fine-tuning.

\begin{table}[t]
\centering
\small
\caption{\textbf{MuSe matches or beats the CUDNN baseline on both domains.} Headline 1B results on code (24B tokens) and scientific PDFs (24B tokens). Cross-entropy loss, lower is better. $^\dagger$Fine-tuned with CUDNN attention for 0.1\% of pretraining tokens.}
\label{tab:headline}
\begin{tabular}{llcc}
\toprule
\multirow{2}{*}{Domain} & \multirow{2}{*}{Train} & \multicolumn{2}{c}{Test Attention} \\
\cmidrule(lr){3-4}
& & CUDNN & \ours{} \\
\midrule
Code & CUDNN & 0.7026 & --- \\
Code & \ours{} & 0.7108 & 0.7001 \\
Code & \ours{}$^\dagger$ & \textbf{0.6994} & 0.7020 \\
\midrule
PDF & CUDNN & 1.6201 & --- \\
PDF & \ours{} & \textbf{1.6166} & 1.6188 \\
\bottomrule
\end{tabular}
\end{table}

\begin{figure}[t]
\centering
\includegraphics[width=0.71\columnwidth]{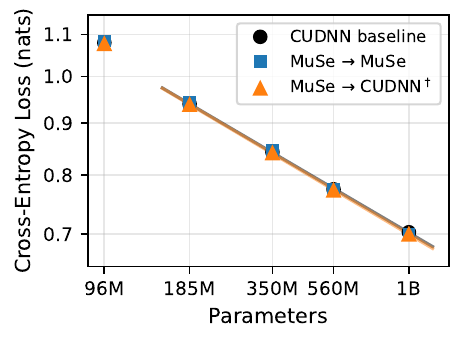}
\caption{Scaling behavior from 96M to 1B parameters. Power law fits (lines) span 185M--1B, where scaling is cleanest; the 96M point falls slightly above the fit. \ours{}-trained models evaluated with CUDNN attention (orange triangles, $\dagger$) track the exact attention baseline, confirming that the approximation preserves scaling properties. $\dagger$The 1B point uses the fine-tuned value after 0.1\% additional CUDNN training.}
\label{fig:scaling}
\end{figure}

\paragraph{Speedup Analysis}

At 1B scale on a single node of 8 A100 GPUs with 64k context, \ours{} achieves 88.9 KTok/s compared to 65.3 KTok/s for CUDNN Flash Attention---a \textbf{36\% throughput improvement}.
Our Pallas kernels are not highly optimized relative to production-quality CUDA; substantially larger speedups are achievable given the 64$\times$ far-field sparsity.
Combined with the minimal quality degradation shown above, this demonstrates that \ours{} provides substantial practical speedups for long-context pretraining.

Figure~\ref{fig:training_curves} shows training loss versus wall-clock time on both code and scientific PDF domains.
At any given loss level, \ours{} reaches that point faster than CUDNN, with the horizontal gap representing the throughput advantage.
The curves are nearly identical when plotted against tokens (see Appendix), confirming that \ours{} achieves equivalent sample efficiency---the speedup comes purely from faster iteration, not from any change in learning dynamics.

\begin{figure*}[t]
\centering
\includegraphics[width=0.855\textwidth]{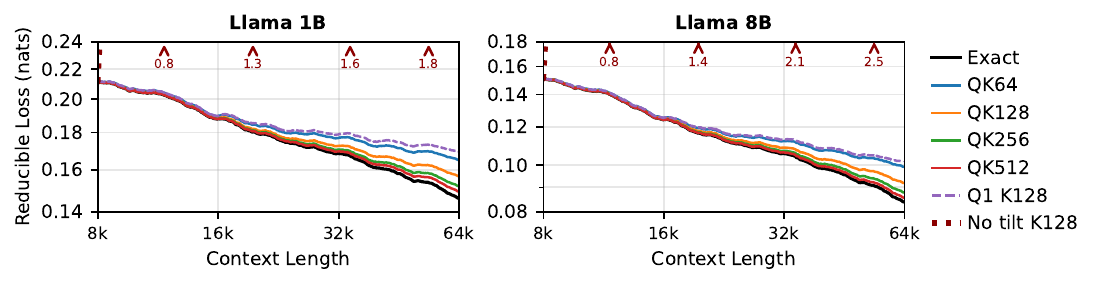}
\caption{Cumulative mean loss versus context length on Project Gutenberg text for Llama 3.2-1B (left) and Llama 3.1-8B (right).
\ours{} with various cluster counts (QK64--QK512) tracks exact attention closely, with the gap reducing by ${\sim}1.8\times$ per doubling for 1B and ${\sim}2\times$ for 8B.
Without exponential tilting (``No tilt K128'', dotted), loss explodes immediately.
Query clustering provides additional benefit: QK64 outperforms Q1~K128 despite using half the key clusters, demonstrating that clustering queries more than doubles practical quality.
Y-axes show reducible loss (cumulative mean minus fitted irreducible loss: 2.30 nats for 1B, 1.96 nats for 8B).}
\label{fig:llama_loss_vs_position}
\end{figure*}

\paragraph{Where the Time Goes}
Table~\ref{tab:runtime_breakdown} breaks attention runtime down by component at our 1B operating point (QK128R8, 64k context).
Retrieval and the block-diagonal exact attention together account for two-thirds of the cost, while monopole summarization and clustering, the two components central to our novel use of query-cluster-specialized exponential tilting, are minor contributors at $8.8\%$ and $8.4\%$.
The dominant cost is therefore retrieval rather than the approximation machinery itself, making retrieval the natural target for further kernel optimization.
Clustering shares the $O(NCD)$ scaling of the monopole stage (Appendix~\ref{app:clustering}), so its share stays roughly constant as cluster count grows and it cannot become a bottleneck.

\begin{table}[t]
\centering
\small
\caption{Attention runtime by component at the 1B operating point (QK128R8, 64k context, head dimension 64). Retrieval dominates; monopole summarization and clustering, central to \ours{}'s query-cluster-specialized tilting, together account for only ${\sim}17\%$.}
\label{tab:runtime_breakdown}
\begin{tabular}{lc}
\toprule
Component & Share of runtime \\
\midrule
Retrieval                      & 37\% \\
Block-diagonal exact attention & 30\% \\
XLA fusions, sorts, transposes & 18\% \\
Monopole summarization         & 8.8\% \\
Clustering                     & 8.4\% \\
\bottomrule
\end{tabular}
\end{table}

\paragraph{Scaling Analysis}

Table~\ref{tab:scaling_sg} and Figure~\ref{fig:scaling} present scaling results on the code domain from 96M to 1B parameters.
We train each model for approximately 20 tokens per parameter, following compute-optimal scaling~\citep{hoffmann2022_chinchilla}.
We report cross-entropy loss for models trained with either CUDNN Flash Attention or \ours{}, evaluated with both attention methods to assess interchangeability.
At scales up to 560M, \ours{}-trained models match or exceed the CUDNN baseline when evaluated with CUDNN attention.
At 1B scale, \ours{}-trained models tested with \ours{} attention beat the baseline (0.7001 vs 0.7026), though there is minor adaptation when tested with CUDNN attention (0.7108); this adaptation is removed with brief CUDNN fine-tuning (0.1\% of pretraining tokens), after which the model beats the baseline even when tested with CUDNN attention (0.6994 vs 0.7026).
Power law fits to the 185M--1B data (excluding the undertrained 96M point) show that \ours{} follows the same scaling law as exact attention (loss $\propto$ params$^{-0.17}$, $R^2 > 0.999$).

\begin{table}[t]
\centering
\small
\caption{\textbf{MuSe matches baseline scaling laws across five model sizes.} Scaling results on code domain (cross-entropy loss). Lower is better. $^\dagger$Fine-tuned with CUDNN attention for 0.1\% of pretraining tokens.}
\label{tab:scaling_sg}
\begin{tabular}{llccc}
\toprule
\multirow{2}{*}{Params} & \multirow{2}{*}{Train} & \multicolumn{2}{c}{Test Attention} \\
\cmidrule(lr){3-4}
& & CUDNN & \ours{} \\
\midrule
1B & CUDNN & 0.7026 & --- \\
1B & \ours{} & 0.7108 & 0.7001 \\
1B & \ours{}$^\dagger$ & \textbf{0.6994} & 0.7020 \\
\midrule
560M & CUDNN & 0.7745 & --- \\
560M & \ours{} & \textbf{0.7728} & 0.7746 \\
\midrule
350M & CUDNN & 0.8427 & --- \\
350M & \ours{} & \textbf{0.8407} & 0.8453 \\
\midrule
185M & CUDNN & 0.9400 & --- \\
185M & \ours{} & \textbf{0.9384} & 0.9435 \\
\midrule
96M & CUDNN & 1.080 & --- \\
96M & \ours{} & \textbf{1.077} & 1.084 \\
\bottomrule
\end{tabular}
\end{table}

\paragraph{Method Comparison}

Table~\ref{tab:ablation} compares \ours{} against MoBA at 185M and 560M scale with matched sparsity (8k block diagonal, 64$\times$ far-field sparsity).
MoBA uses spatial block structure without semantic clustering, retrieving one 512-token block in the far field per query to achieve the matched sparsity.

\ours{} is the only method that \emph{beats} the exact CUDNN baseline at both scales (0.9384 vs 0.9400 at 185M; 0.7728 vs 0.7745 at 560M), demonstrating that our approximation can act as a beneficial regularizer.
MoBA performs significantly worse at both scales (0.9714 at 185M, 0.7958 at 560M), showing that semantic clustering---not just sparsity---is critical for approximation quality.
Even if MoBA reallocates compute from the block diagonal to reduce far-field sparsity to 8$\times$, \ours{} maintains an advantage (Appendix~\ref{app:moba_comparison}).
The importance of query clustering is discussed further in Section~\ref{sec:generalization}.

\begin{table}[t]
\centering
\small
\caption{\textbf{MuSe beats the exact baseline; MoBA does not.} Method comparison at 185M and 560M scale (cross-entropy loss). All methods use matched sparsity (8k block diagonal, 64$\times$ far-field sparsity). Lower is better.}
\label{tab:ablation}
\begin{tabular}{lcc}
\toprule
\multirow{2}{*}{Method} & \multicolumn{2}{c}{Test Attention} \\
\cmidrule(lr){2-3}
& CUDNN & Method \\
\midrule
\multicolumn{3}{l}{\textit{185M:}} \\
CUDNN (baseline) & 0.9400 & --- \\
\ours{} & \textbf{0.9384} & 0.9435 \\
MoBA & 0.9714 & 1.0027 \\
\midrule
\multicolumn{3}{l}{\textit{560M:}} \\
CUDNN (baseline) & 0.7745 & --- \\
\ours{} & \textbf{0.7728} & 0.7746 \\
MoBA & 0.7958 & 0.8209 \\
\bottomrule
\end{tabular}
\end{table}

\paragraph{Head Dimension}

Following recent open-weight releases from major labs \citep{openai2025_gptoss}, our main experiments use head dimension 64.
Table~\ref{tab:head128} validates that \ours{} also works well with head dimension 128, as used by Llama models.
At 185M scale, \ours{} with head dimension 128 slightly outperforms the CUDNN baseline (0.9113 vs 0.9121) and shows strong interchangeability.
Microbenchmarks confirm similar approximation quality: at our operating point (QK128R8), head dimension 128 achieves 0.0100 relative squared error and 0.994 correlation, compared to 0.0094 and 0.995 for head dimension 64.

\begin{table}[h]
\centering
\small
\caption{\textbf{MuSe generalizes to head dimension 128.} Head dimension 128 comparison at 185M scale (cross-entropy loss).}
\label{tab:head128}
\begin{tabular}{lcc}
\toprule
\multirow{2}{*}{Train} & \multicolumn{2}{c}{Test Attention} \\
\cmidrule(lr){2-3}
& CUDNN & \ours{} \\
\midrule
CUDNN & 0.9121 & 0.9267 \\
\ours{} & \textbf{0.9113} & 0.9159 \\
\bottomrule
\end{tabular}
\end{table}

\subsection{Generalization and Long-Context Validation}
\label{sec:generalization}

%
%

\paragraph{Evaluation on Existing Models}

To validate that \ours{} generalizes beyond models trained with it, we evaluate on pretrained Llama 3.2-1B (head dimension 64) and Llama 3.1-8B (head dimension 128) using 64k-token passages from Project Gutenberg.
Figure~\ref{fig:llama_loss_vs_position} shows cumulative mean loss as a function of context length.
With tilting enabled, all \ours{} configurations track exact attention with modest gaps that decrease with cluster count.
The gap approximately halves with each doubling of QK clusters, and QK512 nearly matches exact attention for both models.

Critically, without exponential tilting by query centroids (``No tilt K128''), loss explodes immediately---the approximation fails catastrophically on pretrained models.
This demonstrates that tilting is essential for drop-in compatibility: attention heads with non-trivial mean query values cannot be approximated without accounting for this bias.

Query clustering provides substantial additional benefit beyond tilting alone.
Comparing Q1~K128 (single query cluster with tilting) to QK64 (64 query and key clusters), QK64 achieves lower error despite using half as many key clusters.
This shows that clustering queries more than doubles the effective quality of the approximation.
This benefit comes at negligible cost: attention with few queries is memory-bound, with operational intensity approximately proportional to the query count~\citep{ye2025_flashinfer}, so tilting by moderate numbers of centroids (32--64) is essentially free on current hardware.

\paragraph{Context Utilization During Training}

\begin{figure}[t]
\centering
\includegraphics[width=0.88\columnwidth]{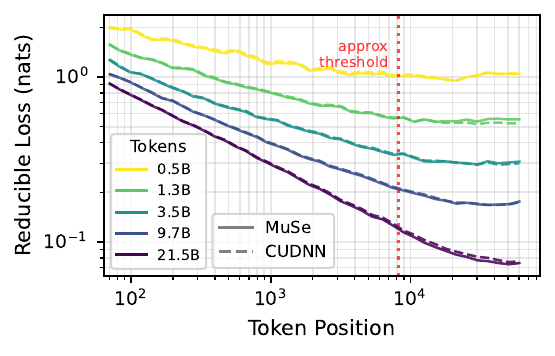}
\caption{Loss versus position during 1B model training on code data.
Curves show progression from early training (light) to late training (dark) at 5 checkpoints spanning 0.5B to 21.5B tokens.
\ours{} (solid) closely tracks cuDNN attention (dashed) throughout training.
The vertical line marks $2^{13}=8192$ tokens, beyond which \ours{} uses far-field approximation.
Both methods learn to utilize the full 64k context, with loss continuing to decrease even in the approximated region.}
\label{fig:loss_vs_position_training}
\end{figure}

Figure~\ref{fig:loss_vs_position_training} demonstrates that models trained with \ours{} learn to utilize the full 64k context throughout training.
Loss continues to decrease with position even beyond the approximation threshold at $2^{13}$ tokens, where \ours{} transitions from exact to far-field computation.
The close correspondence between \ours{} and exact attention at all training stages confirms that the approximation does not impair the model's ability to learn long-range dependencies.

\paragraph{Long-Context Retrieval Evaluation}
\label{sec:retrieval_eval}
A commonly used proxy for long-context downstream capability is retrieval, probed by standardized evaluations such as RULER~\citep{hsieh2024ruler}.
Since \ours{} is used only during training and replaced by exact attention at inference, such evaluations do not test \ours{} attention directly; they test whether a model trained with \ours{} has learned to use the retrieval capability of exact attention once it is restored.
A strong result is therefore a further demonstration of transferability between \ours{} and exact attention, compounding the pretrained-model (Llama) generalization shown above.
RULER, however, targets powerful generalist instruction-tuned models rather than the 1B-scale, domain-specific base models we train.
Both \ours{}- and CUDNN-trained models therefore score poorly on most RULER tasks, the code models in particular near zero on everything but basic single-needle retrieval (needle-in-a-haystack, NIAH); we read this as reflecting domain shift and limited instruction-following far more than a retrieval deficit.
Per-task results are reported in Table~\ref{tab:ruler}.

Rather than attempt to disentangle the contributions of this domain shift and limited instruction-following capability to the complex-variant RULER scores, we designed a custom NIAH evaluation that is in-domain and natural for our code models, a domain where retrieval is intrinsic: function definitions are retrieved via imports across long distances, variable bindings span large scopes, and cross-file dependencies are pervasive.
It uses cross-file function imports as the retrieval primitive: a file defining a function (the \emph{key}) is placed in a repository, and a separate file imports and calls it (the \emph{query}); the function name carries a random uid (the \emph{value}), so correct completion requires retrieval rather than guessing.
We construct single-needle, multi-key ($K \in \{2,4\}$ distractor definitions), and multi-query ($N \in \{2,4\}$ needles probed at different depths) variants.
Table~\ref{tab:custom_niah} reports uid-level exact match aggregated over needle depth: \ours{}-trained models match or exceed exact-attention-trained models on every variant.
Resolving by depth (Appendix~\ref{app:niah_depth}), \ours{}-trained accuracy remains above 89\% in every depth bucket, whereas the exact-attention baseline degrades with depth on the harder variants (\eg 64--97\% across depth buckets on multi-key $K{=}4$).
\begin{table}[t]
\centering
\small
\caption{\textbf{\ours{}-trained models match or exceed exact-attention-trained models on in-distribution retrieval.} Custom code NIAH at 64k context: uid-level exact match ($\pm$ binomial std.\ error). Column headers denote the training regime; all evaluation uses exact attention.}
\label{tab:custom_niah}
\begin{tabular}{lrcc}
\toprule
Task & $n$ & CUDNN & \ours{} \\
\midrule
Single-needle    & 2547 & 97.80 $\pm$ 0.29\% & \textbf{98.35 $\pm$ 0.25\%} \\
Multi-key $2\times$    & 2475 & 92.73 $\pm$ 0.52\% & \textbf{98.46 $\pm$ 0.25\%} \\
Multi-key $4\times$    & 2421 & 81.50 $\pm$ 0.79\% & \textbf{97.36 $\pm$ 0.33\%} \\
Multi-query $2\times$  &  538 & 88.48 $\pm$ 1.38\% & \textbf{98.70 $\pm$ 0.49\%} \\
Multi-query $4\times$  & 1060 & 83.58 $\pm$ 1.14\% & \textbf{98.77 $\pm$ 0.34\%} \\
\bottomrule
\end{tabular}
\end{table}

\paragraph{Downstream Evaluation}
\label{sec:downstream_eval}
As a holistic, real-task long-context benchmark we evaluate the code-completion subset of LongBench~\citep{bai2024longbench} (RepoBench-P and LCC), which is in-distribution for our code models; the remaining LongBench tasks require instruction following.
\ours{} matches exact attention on both subtasks (Table~\ref{tab:longbench_code}), with differences comparable to the reported standard error.

\begin{table}[t]
\centering
\small
\caption{\textbf{\ours{} matches exact attention on LongBench code completion.} Edit similarity ($\pm$ harness-reported std.\ error). Both methods are within the reported standard error of each other.}
\label{tab:longbench_code}
\begin{tabular}{lcc}
\toprule
Task & CUDNN & \ours{} \\
\midrule
RepoBench-P  & 0.382 $\pm$ 0.012 & 0.393 $\pm$ 0.011 \\
LCC          & 0.329 $\pm$ 0.009 & 0.314 $\pm$ 0.010 \\
\bottomrule
\end{tabular}
\end{table}

We additionally evaluate 1B models on ARC-Easy and SciQ with the lm-eval harness.
\ours{}-trained and CUDNN-trained models perform comparably (\eg 48.7\% vs 47.9\% on ARC-Easy), with all differences within one standard error, confirming that the approximation does not impair downstream task performance (full results in Appendix~\ref{app:downstream}).

\section{Conclusion}
\label{sec:conclusion}

%

We have presented Multipole Semantic Attention (\ours{}), an efficient approximation to softmax attention that clusters queries and keys separately in their learned representation spaces.
By computing query-specific cluster summaries and retrieving high-attention clusters, \ours{} achieves high far-field sparsity with under 1\% relative squared error.

Our experiments demonstrate practical benefits at scale: 2$\times$ speedup over CUDNN Flash Attention on isolated attention layers, and 36\% wallclock throughput improvement when pretraining 1B parameter models at 64k context.
Models trained with \ours{} achieve comparable loss to exact attention baselines and remain interchangeable at test time.
Exponential tilting by query centroids is essential for compatibility with pretrained models, and query clustering provides substantial additional quality at negligible cost.

Limitations include the focus on head dimension 64 in most experiments, though preliminary results at dimension 128 are promising.
Future work includes scaling to larger models, kernel optimization, and investigating whether query-dependent tilting benefits other approximate attention mechanisms.
\ours{} applies to grouped-query attention without modification, as exact attention does; since the keys are shared within a group, clustering them once per group rather than once per query head is a natural further optimization.


\pagebreak
\section*{Acknowledgements}
This work was supported by the Hessian research priority programme LOEWE within the project ``WhiteBox'', and the Aleph Alpha Collaboration Lab 1141. It benefited from the Federal Ministry for Research, Technology and Space (BMFTR) project ``XEI: Extremely Efficient Inference for Large Context Length'' (XEI), project identification number 01IS24079B, and from funding by the Deutsche Forschungsgemeinschaft (DFG, German Research Foundation) under Germany's Excellence Strategy -- EXC-3057.
We thank Manuel Brack, Moritz Willig, Felix Friedrich, Felix Divo, Florian Busch, and Rubin H\"arle for helpful discussions and feedback on the manuscript.


\section*{Impact Statement}

Our work on Multipole Semantic Attention (\ours{}) has significant potential for democratizing access to long-context language models by substantially reducing computational costs.
This could enable broader participation from resource-limited researchers and reduce environmental impact, though network effects may also happen.
The increased efficiency in processing longer contexts may benefit applications in scientific research, education, and document analysis.
However, these same capabilities raise concerns about potential misuse in automated disinformation campaigns, enhanced surveillance through efficient processing of large text corpora, and displacement of knowledge workers.
The interchangeability of \ours{}-trained models with standard attention at inference time may complicate accountability and bias auditing efforts.
There is also the risk of over-emphasizing computational efficiency at the expense of other crucial aspects like robustness, factual accuracy, and alignment, necessitating careful monitoring of downstream applications and continued research into behavioral differences emerging from approximate attention mechanisms.


\bibliography{references}
\bibliographystyle{icml2026}


\clearpage
\appendix

\section{Algorithm Pseudocode}
\label{app:algorithm}

\begin{algorithm}[h]
\caption{\ours{} Causal Attention}
\label{alg:muse}
\begin{algorithmic}[1]
\REQUIRE Queries $Q$, Keys $K$, Values $V$ of length $N$; block size $B$; cluster count $C$; retrieval counts $k_1$, $k_2$
\ENSURE Output $O$ of length $N$

\STATE \textbf{// Clustering (global, ignores block boundaries)}
\STATE Cluster $Q$ into $C$ clusters; decompose $q = \qbar_i + \qres$
\STATE Cluster $K$ into $C$ clusters; decompose $k = \kbar_j + \kres$

\STATE \textbf{// Per-block summaries: $C \times C \times (N/B)$ entries}
\FOR{each spatial block $b$, query cluster $i$, key cluster $j$}
    \STATE $\mubar_{ijb}, \kbar_{ijb}, \vbar_{ijb} \gets$ attention-weighted stats of $(K,V)$ in $(j, b)$ from $\qbar_i$
\ENDFOR

\STATE \textbf{// Causal accumulation along spatial axis}
\FOR{each block $b = 1, \ldots, N/B - 1$}
    \STATE $\text{accum}_{ij,b} \gets \text{accum}_{ij,b-1} \oplus \text{summary}_{ij,b-1}$ \COMMENT{logsumexp merge}
\ENDFOR

\STATE \textbf{// Two-level retrieval + attention (per query)}
\FOR{each query $q$ in cluster $i$, block $b$}
    \STATE Select top-$k_1$ key clusters from $\text{accum}_{i,:,b}$ \COMMENT{Cluster selection}
    \STATE Select top-$k_2$ blocks per cluster from $\text{summary}_{i,j,:}$ \COMMENT{Spatial selection}
    \STATE $O_{\text{retrieved}} \gets$ exact attention on $k_1 \times k_2$ (cluster, block) pairs
    \STATE $O_{\text{approx}} \gets$ approximate attention on remaining clusters/blocks
    \STATE $O_{\text{local}} \gets$ exact causal attention within block $b$
\ENDFOR

\STATE \textbf{return} logsumexp-weighted merge of $O_{\text{retrieved}}$, $O_{\text{approx}}$, $O_{\text{local}}$
\end{algorithmic}
\end{algorithm}

\section{Error Analysis of the Far-Field Approximation}
\label{app:dipole}

Here we analyze the error of the monopole approximation via a polynomial expansion of the attention function, and derive the \emph{dipole correction}---an alternative to retrieval that cancels the leading error term.
While retrieval performs better empirically, this analysis pins down what the monopole discards, why K-means is the right clustering objective, and where a cheaper partially specialized dipole loses accuracy.

\paragraph{Polynomial Expansion via Cumulant Generating Functions}
The keys $k$ of some cluster $j$ can be considered as a probability distribution of equally weighted point masses.
They have moment generating function $\mathcal{M}_j(q) := \mathbb{E}_{k \in C_j} \exp(q\cdot k)$
and cumulant generating function (CGF) $\mathcal{K}_j(q) := \ln \mathcal{M}_j(q)$, whose derivatives give the cumulants (mean, variance, skewness, etc.).

Attending to keys with query centroid $\qbar_i$ corresponds to an exponential tilt of this distribution.
Define the CGF of the joint key-value distribution: $\mathcal{K}_{j}(s, t) := \ln \mathbb{E}_{k,v \in C_j} \exp(s\cdot k + t \cdot v)$.
The output of softmax attention on cluster $j$ for query $q$ is:
$V_j(q) = \left.\frac{\partial \mathcal{K}_j(s, t)}{\partial t}\right|_{s=q, t=0}$.

For a query $q = \qbar_i + \qres$, we can approximate $V_j(q)$ by polynomial expansion around $\qbar_i$:
\begin{equation}
\label{eq:dipole-value}
    V_{ij}(\qres) =
    \vbar_{ij} + \Cov_{ij}(v,k) \qres + \frac{1}{2}\qres^T \operatorname{Skew}_{ij}(v,k,k) \qres + \ldots
\end{equation}
where $\vbar_{ij}$, $\Cov_{ij}(v,k)$, and $\operatorname{Skew}_{ij}(v,k,k)$ are cumulants of the exponentially-tilted distribution.

Similarly, the unnormalized attention weight expands as:
\begin{equation}
\label{eq:dipole-mass}
    M_{ij}(\qres) =
    M_j(\qbar_i)
    \exp\left( \qres \cdot \kbar_{ij} + \frac{1}{2}\qres^T \Cov_{ij}(k,k) \qres + \ldots\right)
\end{equation}

\paragraph{Monopole Error and the Clustering Objective}
The monopole approximation can be read directly off \eqref{eq:dipole-value}--\eqref{eq:dipole-mass}: it keeps the value at zeroth order in $\qres$ (the tilted summary $\vbar_{ij}$) and the log-mass at first order ($\mubar_{ij}+\qres\cdot\kbar_{ij}$, Eq.~\eqref{eq:score}), while discarding the linear value term $\Cov_{ij}(v,k)\,\qres$ and the quadratic mass term $\tfrac12\qres^T\Cov_{ij}(k,k)\qres$.
The truncation is asymmetric---value at zeroth order, mass at first---so the leading error is the dropped linear value term $\Cov_{ij}(v,k)\,\qres=\mathcal{O}(\qres)$, equivalently the per-weight bilinear $\mathcal{O}(\qres\cdot\kres)$ of \cref{sec:retrieval}.
This is specifically the \emph{within-cluster} component of the first-order response; the \emph{between-cluster} component---the shifting of attention mass between clusters---is already carried by the score's $\qres\cdot\kbar_{ij}$ dependence (see the Dipole Correction below), so the within-cluster value term is the monopole's only first-order omission.

Averaging the squared value error over the queries of cluster $i$ gives, per key cluster,
\begin{equation}
\label{eq:monopole-error}
\begin{split}
&\mathbb{E}_{\qres}\,\bigl\lVert V_{ij}(\qres)-\vbar_{ij}\bigr\rVert^2\\
&=\operatorname{Tr}\!\bigl(\Cov_{ij}(v,k)\,\Cov(\qres,\qres)\,\Cov_{ij}(k,v)\bigr)\\
&\quad+\mathcal{O}(\qres^3).
\end{split}
\end{equation}
by substituting the linear term of \eqref{eq:dipole-value}, taking the expectation (with $\Cov(\qres,\qres)=\mathbb{E}_{q\in i}[\qres\,\qres^T]$), and applying trace cyclicity; the next-order term is the third moment of the residuals, nonzero for asymmetric clusters.
The aggregate within-cluster term omitted by the monopole is thus $\bar C_i\,\qres$ to leading order, with $\bar C_i=\sum_j p_i(j)\,\Cov_{ij}(v,k)$ and centroid attention weights $p_i(j)$; this merged covariance $\bar C_i$ recurs in the dipole construction below.

The per-cluster error \eqref{eq:monopole-error} is controlled by the clustering objectives:
\begin{equation}
\label{eq:kmeans-bound}
\begin{split}
&\operatorname{Tr}\!\bigl(\Cov(v,k)\,\Cov(\qres,\qres)\,\Cov(k,v)\bigr)\\
&\le \operatorname{Tr}\Cov(\qres,\qres)\,\operatorname{Tr}\Cov(\kres,\kres)\,\operatorname{Tr}\Cov(\tilde v,\tilde v),
\end{split}
\end{equation}
using $\operatorname{Tr}(AB)\le\operatorname{Tr}(A)\operatorname{Tr}(B)$ for positive semidefinite matrices and $\lVert\Cov(v,k)\rVert_F^2\le\operatorname{Tr}\Cov(\tilde v,\tilde v)\,\operatorname{Tr}\Cov(\kres,\kres)$ (Cauchy--Schwarz on the cross-covariance).
The bound \emph{factorizes} into a product of three traces---the query, key, and value within-cluster variances; the first two are exactly the K-means objectives for queries and keys, while $\operatorname{Tr}\Cov(\tilde v,\tilde v)$ is not itself a clustering objective, since we cluster queries and keys rather than values. This product form is what justifies clustering queries and keys independently.\footnote{This decoupling is loose: the exponent-level second moment is $\mathbb{E}[(\qres\cdot\kres)^2]=\operatorname{Tr}(\Cov(\qres,\qres)\Cov(\kres,\kres))$, minimized by clustering keys in the Mahalanobis metric $\Cov(\qres,\qres)$ and queries in $\Cov(\kres,\kres)$---a whitening before Euclidean K-means, using global residual covariances to avoid the query--key fixed point. Euclidean K-means is the isotropic proxy: exact under isotropy, and otherwise differing from the exact objective only by the second-order alignment between residual covariance and cluster geometry. We considered but did not adopt it, as the per-pass whitening overhead is not repaid by the modest accuracy gain.}
One caveat applies throughout: the covariances in these expressions are $\qbar_i$-tilted, whereas K-means minimizes the \emph{untilted} within-cluster variance; the two agree to leading order in the residual magnitudes.

\paragraph{Dipole Correction}
The attention output's linear-in-$\qres$ response splits into a \emph{between-cluster} part (which clusters are reweighted) and a \emph{within-cluster} part (how each cluster summary shifts).
The monopole already realizes the between-cluster part; differentiating its cluster softmax at $\qres=0$,
\begin{equation}
\label{eq:between-cluster}
\begin{split}
&\partial_{\qres}\big|_{\qres=0}\sum_j \softmax_j(\qres\cdot\kbar_{ij}+\mubar_{ij})\,\vbar_{ij}\\
&=\sum_j p_i(j)\,\vbar_{ij}\,(\kbar_{ij}-\bar k^{(i)})^T,
\end{split}
\end{equation}
the $p_i$-weighted cross-covariance between value and key summaries across clusters (acting on $\qres$), with $\bar k^{(i)}=\sum_j p_i(j)\,\kbar_{ij}$.
The dipole supplies the missing within-cluster part by restoring the linear value term while leaving the mass unchanged:
\begin{equation}
\label{eq:dipole-shift}
\vbar_{ij}\;\mapsto\;\vbar_{ij}+\Cov_{ij}(v,k)\,\qres
\qquad(\text{score still }\mubar_{ij}+\qres\cdot\kbar_{ij}).
\end{equation}
This cancels the leading monopole error \eqref{eq:monopole-error}; the residual is the next term of \eqref{eq:dipole-value}, $\tfrac12\qres^T\Skew_{ij}(v,k,k)\qres$, an $\mathcal{O}(\qres^2)$ output error (per-weight $\mathcal{O}(\qres^2\kres^2)$).

Summed over key clusters, the per-query correction is $\bigl(\sum_j p(j\mid q)\,\Cov_{ij}(v,k)\bigr)\qres$.
Replacing the query-dependent weights $p(j\mid q)$ by the query-cluster-constant centroid weights $p_i(j)$ pre-merges it into a single matrix per query cluster, applied once per query as $\bar C_i\,\qres$:
\begin{equation}
\label{eq:premerge}
\bar C_i=\sum_j p_i(j)\,\Cov_{ij}(v,k).
\end{equation}
Crucially, the merge weights $p_i(j)$ are the monopole's own centroid cluster weights, already determined by the summary masses $S_{ij}$ (equivalently $\mubar_{ij}=\log S_{ij}$), so the pre-merge is not merely an efficiency: it is the \emph{correct} composition with the monopole, handling the between-cluster $\qbar\cdot\kbar$ interaction exactly and once.
Moreover $\bar C_i$ is the same aggregate that governs the monopole error \eqref{eq:monopole-error}, so the dipole literally subtracts the leading monopole error.
The substitution is legitimate at the dipole's order: $p(j\mid q)-p_i(j)=\mathcal{O}(\qres)$ multiplies an already-$\mathcal{O}(\qres)$ correction, an $\mathcal{O}(\qres^2)$ term that has already been discarded.

Two costs must be kept separate.
\emph{Consumption} of the merged correction drops from $\mathcal{O}(N\DC D^2)$ (one matrix per cluster) to $\mathcal{O}(N D^2)$ and is tilt-independent; \emph{construction} of the tilted $\Cov_{ij}(v,k)$ costs $\mathcal{O}(\DC N D^2)$, which the next paragraph addresses.

\paragraph{Untilted (Partially Specialized) Covariances}
Because the between-cluster interaction is secured exactly by the (free) merge weights $p_i(j)$, the only quantity left to approximate is the within-cluster covariance itself---and sharing it keeps the penalty's key dependence at within-cluster scale ($\kres$) rather than centroid scale, with the degradation confined to the query side.
We share one covariance per key cluster, \emph{untilted}, $\Cov_{0j}(v,k)$, dropping construction from $\mathcal{O}(\DC N D^2)$ to $\mathcal{O}(N D^2)$.
Since the derivative of the covariance along the tilt is the third cumulant,
\begin{equation}
\label{eq:untilt-expand}
\Cov_{ij}(v,k)=\Cov_{0j}(v,k)+\Skew_j(v,k,k)\cdot\qbar_i+\mathcal{O}(\qbar_i^2),
\end{equation}
substituting $\Cov_{0j}$ injects $\Skew_j(v,k,k)\cdot\qbar_i$ into the coefficient of $\qres$, degrading the value error from $\mathcal{O}(\qres^2\kres^2)$ to $\mathcal{O}(\qbar_i\cdot\qres\cdot\kres^2)$---the same third-cumulant tensor as the dipole residual, with one residual factor $\qres$ replaced by the full centroid $\qbar_i$.
The key side ($\kres^2$) is unchanged; the degradation is purely query-side, residual $\to$ centroid.

Centering the shared tilt at the mass-weighted mean query centroid $\qbar_\star$ rather than at $0$ keeps the $\mathcal{O}(N D^2)$ cost (a single tilt point) and replaces the bound by $\mathcal{O}(\lvert\qbar_i-\qbar_\star\rvert\cdot\qres\cdot\kres^2)$.
The penalty is now governed by the spread of the query centroids about their mean---a distinct quantity from the mean-query magnitude $\lvert\qbar_\star\rvert$ that drives the need for tilting in the first place (Figure~\ref{fig:llama_loss_vs_position}).
A single shared tilt point can be exact for at most one query cluster, so the residual penalty is the spread of the remaining centroids about it, vanishing only when the centroids coincide---the price of one shared covariance in place of one per query cluster.\footnote{Collapsing the per-query-cluster weights as well---an occupancy-weighted or globally shared covariance---would forfeit the between-cluster interaction those weights encode and degrade the error back to centroid scale, so the weights are retained.}

We use retrieval rather than a dipole in our main method: retrieval achieves a larger error reduction (see below), and combining a dipole with retrieval requires subtracting the retrieved clusters' contribution from the merged correction (next paragraph).

\paragraph{Compatibility with Retrieval}
One might hope to combine dipole corrections with retrieval for further accuracy gains.
However, the merged correction $\bar C_i=\sum_j p_i(j)\,\Cov_{ij}(v,k)$ already sums over all key clusters, so when retrieval selects specific clusters for exact attention those clusters have already contributed to $\bar C_i$.
Correctly combining the two would require subtracting the retrieved clusters' contribution from $\bar C_i$ before adding the exact attention---possible, but complex.
One could instead keep the per-$(i,j)$ dipoles un-merged and mask the retrieved clusters exactly as the monopole path does, but this forfeits the $\mathcal{O}(N D^2)$ consumption that motivated the merge.
Given that retrieval alone already provides larger accuracy gains than dipoles, we opted not to pursue this combination.

\paragraph{Empirical Comparison with Retrieval}
In practice, we found that retrieval of high-attention clusters provides better accuracy than dipole corrections.
At K128, monopole-only approximation achieves 0.0227 relative squared error; adding the untilted (partially specialized) dipole corrections reduces this to 0.0150 (1.5$\times$ improvement), while adding retrieval instead reduces it to 0.00622 (3.6$\times$ improvement).
We therefore use retrieval in our main experiments.

\section{Clustering Details}
\label{app:clustering}

We use streaming K-means to obtain cluster centroids, followed by a final assignment pass.

\paragraph{Centroid Computation}
Centroids are computed globally across the entire sequence.
Given $N$ input vectors (queries or keys), we compute $K$ cluster centroids as follows:
\begin{enumerate}
    \item \textbf{Initialization:} Shuffle inputs with a fixed random seed, then uniformly subsample every $\lfloor N/K \rfloor$-th vector to obtain $K$ initial centroids.
    \item \textbf{Streaming update:} Process all $N$ vectors in a single pass using minibatches of 64 vectors. For each minibatch, compute similarities to current centroids using tensor cores, then assign each vector to its nearest centroid. For each assigned vector $x$, update the centroid total and count as $t \leftarrow \beta \cdot t + x$ and $c \leftarrow \beta \cdot c + 1$ (with $\beta = 0.9$), processing assignments sequentially within the minibatch. The centroid is recovered as $t/c$.
\end{enumerate}
A single pass suffices because the effective sample size per cluster grows with sequence length; at 64k tokens with 128 clusters, each cluster sees $\sim$500 vectors on average.
All clustering computations use float16 precision for efficiency.

\paragraph{Balanced Assignment}
After computing global centroids, assignment is performed independently within each spatial block.
Each vector is assigned to its nearest centroid subject to a maximum cluster size of $4 \times B/K$ per block, where $B$ is the spatial block size.
This static cap simplifies prototyping with pure JAX implementations; the final Pallas kernels could handle variable cluster sizes but we retain the cap for consistency.
The $4\times$ average factor was chosen empirically as it rarely binds in practice.
Squared distances are computed efficiently via the identity $\|x - c\|^2 = \|x\|^2 - 2\langle x, c \rangle + \|c\|^2$, where $\|x\|^2$ terms cancel when comparing distances to different centroids.

\paragraph{Centroid Gradient Flow}
When computing attention from query centroids (for exponential tilting), we stop gradients from flowing back through the centroid to its constituent queries.
This prevents queries from being trained to ``cluster well,'' a property that provides no benefit when switching to exact attention at test time.
We found this reduces adaptation effects.

\paragraph{Aggregation Weight Gradient Flow}
The initial summary step aggregates keys and values using softmax weights tilted by the query centroid $\qbar$.
As an optional alternative to the brief fine-tuning of Appendix~\ref{app:finetuning}, we treat these tilted aggregation weights as non-differentiable with respect to both $\qbar$ and the keys.
This does not destroy gradient signal to the keys: they are still trained through the derivative of the loss with respect to the \emph{aggregated} keys (the key centroids that enter the summary).
The removed term is specific to \emph{tilted} aggregation of keys/values, via the dependence of that tilting effect on the keys themselves.
Empirically, stopping this gradient slows learning slightly early in training but converges to better non-adapted minima in larger-scale runs: on the 1B code model it removes the adaptation gap entirely, beating the baseline when evaluated with exact attention and without any fine-tuning (Table~\ref{tab:gradfix}).

\begin{table}[h]
\centering
\small
\caption{Effect of treating the tilted aggregation weights as non-differentiable, on the 1B code model evaluated with exact (CUDNN) attention. The non-differentiable-aggregation variant removes the adaptation gap with no fine-tuning.}
\label{tab:gradfix}
\begin{tabular}{lc}
\toprule
Model (evaluated with CUDNN attention) & Loss \\
\midrule
CUDNN baseline                          & 0.7026 \\
\ours{}, default aggregation            & 0.7108 \\
\ours{}, default aggregation $+$ fine-tune & 0.6994 \\
\ours{}, non-differentiable aggregation & \textbf{0.6991} \\
\bottomrule
\end{tabular}
\end{table}

\paragraph{Clustering Cost}
The cost of clustering is dominated by computing the similarity of all $N$ tokens against the $C$ centroids, a matmul of size $N \times C \times D$.
Clustering performs four such matmuls, compared with five in the monopole forward pass and twelve in its backward pass, so clustering is under $20\%$ of the aforementioned computation alone, and a smaller fraction still once retrieval and block-diagonal attention are counted (Table~\ref{tab:runtime_breakdown}).
Both clustering and the monopole stage scale as $O(NCD)$, strictly cheaper than the $O(N^2 D)$ exact attention they replace; since the ratio between clustering and monopole cost is fixed in $N$ and $C$, clustering cannot overtake the rest of the approximation as either grows.
Consistent with this, the clustering share of total runtime remains at $7$--$8\%$ as the cluster count doubles from 64 to 128 to 256 at constant retrieval fraction.

\paragraph{Behaviour Early in Training}
\ours{} is applied from the first training step, with no warmup.
Early in training the learned representations are still forming, yet K-means recovers substantial cluster structure from them, enough for the resulting partitions to be useful.
The higher-entropy attention patterns characteristic of early training are moreover favourable for the monopole component, since smoother attention is captured more accurately by a cluster summary.
Together these let \ours{} track exact attention from the start (Figure~\ref{fig:loss_vs_position_training}).

\section{Hyperparameters}
\label{app:hyperparameters}

\paragraph{Model Architecture}
We use a standard decoder-only transformer with pre-norm (layer normalization before attention and FFN), RoPE position encodings with base frequency $10^5$, and embedding tying.
Model sizes are parameterized by a scale factor $S \in \{3, 4, 5, 6, 8\}$ corresponding to \{96M, 185M, 350M, 560M, 1B\} parameters:
\begin{itemize}
    \item Embedding dimension: $256 \times S$
    \item Attention heads: $4 \times S$ (head dimension 64)
    \item FFN dimension: $1024 \times S$
    \item Layers: 10, 12, 16, 18, 20 for scales 3--8 ($\approx 3S$, rounded to even for layer pairing; scale 8 uses 20 rather than 24 to hit the target parameter count)
    \item Vocabulary size: 32768
    \item Context length: 64k tokens
\end{itemize}
Layers are grouped into super blocks of 2: one local attention layer (sliding window of 256 tokens) followed by one global attention layer.
FFN uses ReLU activation.
Layer normalization uses $\epsilon = 10^{-6}$.

\paragraph{Optimizer}
We use AdamW with $\beta_1 = 0.9$, $\beta_2 = 0.95$, $\epsilon = 10^{-6}$, and weight decay 0.01.
Gradients are clipped to global norm 1.0.
Learning rate follows a warmup-cosine schedule with 4000 warmup steps and final value 10\% of peak.
Peak learning rate is $3 \times 10^{-3}$, scaled by $S^{-0.5}$.

\paragraph{Training Data}
We train on two datasets:
\begin{itemize}
    \item \textbf{Code:} The Stack v2~\citep{lozhkov2024starcoder2stackv2} (\texttt{bigcode/the-stack-v2-train-smol-ids}). We filter to Python files, excluding vendor and generated code. Repositories are bucketed by estimated token count (Python bytes $/$ 3.5) on a log$_2$ scale; we use buckets 5--8 corresponding to 16k--256k tokens per repository. Files within each repository are concatenated with \texttt{<reponame>} and \texttt{<filename>} delimiters, sorted by path depth. Total: 21.5B tokens.
    \item \textbf{Scientific documents:} OLMo 3 LongMino pool~\citep{olmo2025olmo3} (\texttt{allenai/dolma3\_longmino\_pool}), \texttt{science\_tech} topic, bucket \texttt{2e16} (64k--128k tokens per document). Total: 37.5B tokens.
\end{itemize}
Both datasets use separate BPE tokenizers with 32768 vocabulary trained on domain-specific samples.
Training token counts scale with model size: 2.0B, 4.0B, 7.5B, 12.9B, and 23.6B tokens for scales 3--8 respectively.
Since the code corpus totals 21.5B tokens, the largest code run (scale 8) slightly exceeds one epoch (${\sim}1.1\times$); all other runs remain within a single epoch of their corpus.
Batch size is 2 sequences of 64k tokens (128k tokens per step).

\paragraph{\ours{} Operating Point}
For pretraining experiments, we use 128 query clusters, 128 key clusters, top-8 cluster retrieval, top-1 spatial block retrieval, and 8k spatial blocks (Q128K128R8SP1B8k).
This configuration achieves $<$1\% relative squared error on microbenchmarks while providing 2$\times$ speedup over CUDNN Flash Attention.

\paragraph{Hardware}
All pretraining experiments use 8$\times$ NVIDIA A100 (80GB) GPUs in a $(2, 4)$ mesh with data parallelism over the batch dimension and model parallelism over attention heads.
Training uses mixed precision (bfloat16/float32).

\paragraph{Llama Evaluation}
For generalization experiments (Section~\ref{sec:generalization}), we evaluate on pretrained Llama 3.2 1B and Llama 3.1 8B models converted to Flax.
We measure per-position cross-entropy loss on PG-19~\citep{rae2020_gutenberg_pg19} (test split) with 64k context, using 100 document chunks.
Cumulative mean loss is computed across positions and plotted as reducible loss (subtracting fitted irreducible loss).

\paragraph{Approximation Quality Metric}
We report relative squared error (RSE) as our primary measure of approximation quality: $\text{RSE} = \|o_{\text{approx}} - o_{\text{exact}}\|^2 / \|o_{\text{exact}}\|^2$, where $o$ denotes the attention output, averaged across all queries in the evaluation batch.

\section{96M Model Microbenchmarks}
\label{app:microbenchmarks_96m}

This appendix presents detailed microbenchmark results from a 96M parameter model.
We evaluate approximation quality on queries, keys, and values extracted from the trained model with 64k context.
We use 2 sequences across all 60 attention heads (from all layers), with head dimension 64.
Our base configuration uses 128 query clusters, 128 key clusters, top-8 cluster retrieval, top-1 spatial block retrieval, and 8k spatial blocks (Q128K128R8SP1B8k).

\subsection{Speedup vs Exact Attention}
Table~\ref{tab:micro_speedup} compares \ours{} to exact flash attention implementations.
At our base configuration (QK128R8), \ours{} achieves 2$\times$ speedup over CUDNN Flash Attention with 2.6\% relative squared error.

\begin{table}[h]
\centering
\small
\caption{Runtime comparison with exact attention (64k context, head dim 64, 96M model).}
\label{tab:micro_speedup}
\begin{tabular}{lcccc}
\toprule
Method & Time (ms) & Speedup & RSE \\
\midrule
CUDNN Flash & 212.1 & 1.0$\times$ & 0 (exact) \\
Pallas Flash & 243.6 & 0.87$\times$ & 0 (exact) \\
\ours{} QK64R8 & 118.1 & 1.80$\times$ & 0.0315 \\
\ours{} QK128R8 & 105.5 & 2.01$\times$ & 0.0259 \\
\ours{} QK256R8 & 123.3 & 1.72$\times$ & 0.0206 \\
\bottomrule
\end{tabular}
\end{table}

\subsection{Effect of Query Clustering}
Table~\ref{tab:micro_query} shows the effect of varying the number of query clusters while holding other parameters fixed.
Reducing from 128 to 1 query cluster (mathematically equivalent to no query clustering) increases relative squared error by 1.9$\times$ (0.026 to 0.049) with negligible runtime difference, validating the importance of query clustering.

\begin{table}[h]
\centering
\small
\caption{Effect of query cluster count (K128R8SP1B8k fixed). Q1 uses zeroed centroids, mathematically equivalent to no query clustering.}
\label{tab:micro_query}
\begin{tabular}{cccc}
\toprule
Q & RSE & Correlation & Time (ms) \\
\midrule
256 & 0.0214 & 0.9875 & 122.6 \\
128 & 0.0259 & 0.9848 & 105.5 \\
64 & 0.0301 & 0.9825 & 104.3 \\
32 & 0.0338 & 0.9764 & 102.0 \\
16 & 0.0376 & 0.9737 & 103.6 \\
1* & 0.0494 & 0.9685 & 105.6 \\
\bottomrule
\end{tabular}
\end{table}

\subsection{Effect of Key Clustering}
Table~\ref{tab:micro_key} shows the effect of varying key clusters with retrieval fraction held constant (R/K = 1/16).
More clusters with proportionally more retrieval improves accuracy, with K128R8 providing a good speed/accuracy tradeoff.

\begin{table}[h]
\centering
\small
\caption{Effect of key cluster count with fixed retrieval fraction (Q128SP1B8k fixed, R/K = 1/16).}
\label{tab:micro_key}
\begin{tabular}{ccccc}
\toprule
K & R & RSE & Correlation & Time (ms) \\
\midrule
256 & 16 & 0.0180 & 0.9882 & 137.1 \\
128 & 8 & 0.0259 & 0.9848 & 105.5 \\
64 & 4 & 0.0359 & 0.9762 & 106.7 \\
32 & 2 & 0.0504 & 0.9639 & 117.8 \\
16 & 1 & 0.0692 & 0.9539 & 126.1 \\
\bottomrule
\end{tabular}
\end{table}

\subsection{Joint Query and Key Clustering}
Table~\ref{tab:micro_joint} compares performance with and without query clustering as the number of key clusters varies.
Query clustering provides 1.4--2.4$\times$ error reduction, with larger benefits at higher cluster counts where query-specific summaries matter more.

\begin{table}[h]
\centering
\small
\caption{Effect of query clustering across key cluster counts (R/K = 1/16, SP1B8k fixed). ``No Q'' uses zeroed query centroids.}
\label{tab:micro_joint}
\begin{tabular}{ccccccc}
\toprule
\multirow{2}{*}{K} & \multirow{2}{*}{R} & \multicolumn{2}{c}{RSE} & \multicolumn{2}{c}{Corr} & \multirow{2}{*}{Gain} \\
\cmidrule(lr){3-4} \cmidrule(lr){5-6}
& & No Q & With Q & No Q & With Q & \\
\midrule
256 & 16 & 0.0365 & 0.0152 & 0.9749 & 0.9891 & 2.4$\times$ \\
128 & 8 & 0.0494 & 0.0259 & 0.9685 & 0.9848 & 1.9$\times$ \\
64 & 4 & 0.0676 & 0.0414 & 0.9541 & 0.9723 & 1.6$\times$ \\
32 & 2 & 0.0943 & 0.0640 & 0.9400 & 0.9576 & 1.5$\times$ \\
16 & 1 & 0.1295 & 0.0933 & 0.9177 & 0.9373 & 1.4$\times$ \\
\bottomrule
\end{tabular}
\end{table}

\subsection{Monopole Only (No Retrieval)}
Table~\ref{tab:micro_monopole} shows performance without retrieval, comparing monopole approximation with and without query clustering.
Query clustering provides 1.9--2.7$\times$ error reduction for monopole-only---larger than the 1.4--2.4$\times$ benefit observed with retrieval (Table~\ref{tab:micro_joint}).
This confirms that query clustering is most critical for the monopole component; retrieval partially masks the benefit by handling high-attention clusters exactly.

\begin{table}[h]
\centering
\small
\caption{Monopole only (no retrieval), with and without query clustering. ``No Q'' uses zeroed query centroids.}
\label{tab:micro_monopole}
\begin{tabular}{ccccccc}
\toprule
\multirow{2}{*}{QK} & \multicolumn{2}{c}{RSE} & \multicolumn{2}{c}{Corr} & \multirow{2}{*}{Gain} \\
\cmidrule(lr){2-3} \cmidrule(lr){4-5}
& No Q & With Q & No Q & With Q & \\
\midrule
256 & 0.1560 & 0.0586 & 0.9005 & 0.9606 & 2.7$\times$ \\
128 & 0.1812 & 0.0794 & 0.8879 & 0.9430 & 2.3$\times$ \\
64 & 0.2118 & 0.1025 & 0.8697 & 0.9283 & 2.1$\times$ \\
32 & 0.2362 & 0.1273 & 0.8578 & 0.9131 & 1.9$\times$ \\
\bottomrule
\end{tabular}
\end{table}

\subsection{Cluster Count vs Retrieval Work}
Table~\ref{tab:micro_cluster_retrieval} shows a striking result: increasing cluster count improves accuracy while \emph{reducing} retrieval work.
With R=8 fixed, each doubling of cluster count improves error by 1.2--1.3$\times$ while halving the number of keys retrieved per query.
QK512R8 achieves 2$\times$ better accuracy than QK64R8 while retrieving 8$\times$ fewer keys.
This demonstrates that finer clustering improves monopole quality enough to more than compensate for retrieving from smaller clusters.

Table~\ref{tab:micro_retrieval_count} shows the effect of varying retrieval count with cluster count fixed.
Doubling retrieval from R8 to R16 improves error by 1.27$\times$---comparable to the 1.26$\times$ improvement from doubling cluster count (QK128 to QK256) at fixed R=8.

\begin{table}[h]
\centering
\small
\caption{Effect of cluster count with fixed retrieval count R=8 (SP1B8k fixed). Higher cluster counts improve accuracy despite less retrieval work.}
\label{tab:micro_cluster_retrieval}
\begin{tabular}{cccccc}
\toprule
QK & R & RSE & Correlation & Keys/query & Time (ms) \\
\midrule
512 & 8 & 0.0156 & 0.9908 & 1024 & 226.0 \\
256 & 8 & 0.0206 & 0.9887 & 2048 & 123.3 \\
128 & 8 & 0.0259 & 0.9848 & 4096 & 105.5 \\
64 & 8 & 0.0315 & 0.9789 & 8192 & 118.1 \\
\bottomrule
\end{tabular}
\end{table}

\begin{table}[h]
\centering
\small
\caption{Effect of retrieval count with fixed cluster count (QK128SP1B8k fixed).}
\label{tab:micro_retrieval_count}
\begin{tabular}{ccccc}
\toprule
QK & R & RSE & Correlation & Time (ms) \\
\midrule
128 & 16 & 0.0204 & 0.9851 & 140.8 \\
128 & 8 & 0.0259 & 0.9848 & 105.6 \\
128 & 4 & 0.0355 & 0.9739 & 89.6 \\
\bottomrule
\end{tabular}
\end{table}

\subsection{Retrieval Only (No Monopole)}
Table~\ref{tab:micro_retrieval} shows performance with retrieval only, disabling the monopole contribution.
Query clustering provides only 1.07$\times$ benefit for retrieval selection, compared to 2.3$\times$ for monopole-only (Table~\ref{tab:micro_monopole}).
This confirms that query clustering primarily improves monopole quality; the retrieval mechanism selects reasonable clusters even without query-specific summaries.
Importantly, query clustering does not harm retrieval, making it a pure win.

\begin{table}[h]
\centering
\small
\caption{Retrieval only (no monopole) at QK128R8SP1B8k. ``No Q'' uses zeroed query centroids.}
\label{tab:micro_retrieval}
\begin{tabular}{lccc}
\toprule
Setting & RSE & Correlation & Time (ms) \\
\midrule
With Q & 0.0990 & 0.9645 & 103.4 \\
No Q & 0.1061 & 0.9556 & 103.2 \\
\midrule
Q benefit & \multicolumn{2}{c}{1.07$\times$} & \\
\bottomrule
\end{tabular}
\end{table}

\section{1B Model Microbenchmarks}
\label{app:microbenchmarks_1b}

This appendix presents detailed microbenchmark results from our 1B parameter model (320 heads, head dimension 64, 64k context).
These experiments validate hyperparameter choices and characterize approximation quality at scale.

\subsection{Block Size and Spatial Retrieval Tradeoff}
\label{app:block_size}

Table~\ref{tab:1b_block_size} shows the effect of varying block size $B$ while adjusting spatial retrieval $SP$ to maintain constant retrieval span ($SP \times B = 16k$).
With cluster count and cluster retrieval fixed (QK128R4), this sweep illustrates the fundamental tradeoff between exact local attention and approximate far-field attention.

\begin{table}[h]
\centering
\small
\caption{Effect of block size with constant retrieval span (QK128R4, $SP \times B = 16k$, 1B model, 40 heads, 2 sequences). Larger blocks compute more attention exactly but have higher diagonal cost; smaller blocks approximate more attention but eventually become slow due to accumulation overhead.}
\label{tab:1b_block_size}
\begin{tabular}{lcccc}
\toprule
Config & RSE & Correlation & Time (ms) \\
\midrule
QK128R4SP1B16k & 0.00493 & 0.9974 & 105.6 \\
QK128R4SP2B8k  & 0.00704 & 0.9959 & 75.5 \\
QK128R4SP4B4k  & 0.01578 & 0.9901 & 71.8 \\
QK128R4SP8B2k  & 0.03833 & 0.9767 & 129.4 \\
\bottomrule
\end{tabular}
\end{table}

Accuracy degrades as block size decreases (0.00493 $\to$ 0.03833, a $7.8\times$ increase in error) because more of the attention mass---which is predominantly local---must be routed through the approximation rather than computed exactly in the block diagonal.

Runtime exhibits a U-shape: $B=4\text{k}$--$8\text{k}$ is fastest (${\sim}72$--$76$ms), while both $B=16\text{k}$ (expensive diagonal attention) and $B=2\text{k}$ (expensive accumulation over 32 blocks plus 8 spatial selections per query) are slower.
This demonstrates that block size provides a tunable speed/accuracy tradeoff, with $B=8\text{k}$ offering a good balance for 64k context.

We use $SP=1$ (single spatial block per cluster) for simplicity; $SP=2$ with reduced cluster retrieval ($R=4$) shows similar microbenchmark performance but exhibited more adaptation during pretraining.

\subsection{Semantic vs.\ Spatial Retrieval Breadth}
\label{app:rsp_tradeoff}

The far-field retrieval budget is the number of (cluster, spatial-block) pairs attended exactly per query, $k_1 \cdot k_2$, where $k_1$ (denoted $R$) is the number of retrieved clusters and $k_2$ (denoted $SP$) the number of spatial blocks retrieved within each cluster.
A single retrieval count $k = k_1 \cdot k_2$ over the joint space would suffice if quality depended only on the product; separating $k_1$ and $k_2$ is justified only if the split itself matters.
Table~\ref{tab:rsp_tradeoff} varies the split at a fixed budget of eight pairs (QK128, $B{=}8\text{k}$): shifting budget from semantic breadth ($k_1$) toward spatial breadth ($k_2$) degrades approximation quality, with relative squared error nearly tripling from the semantic-heavy R8SP1 (our operating point) to the spatial-heavy R2SP4.
The R4SP2 configuration coincides with the $SP2B8k$ row of Table~\ref{tab:1b_block_size}.

\begin{table}[h]
\centering
\small
\caption{Semantic vs.\ spatial retrieval breadth at a fixed far-field budget of $k_1 \cdot k_2 = 8$ pairs per query (1B model, QK128, $B{=}8\text{k}$, 64k context, 40 heads). Semantic breadth ($k_1$) dominates spatial breadth ($k_2$) on our data.}
\label{tab:rsp_tradeoff}
\begin{tabular}{lcccc}
\toprule
Config & $k_1$ ($R$) & $k_2$ ($SP$) & RSE & Correlation \\
\midrule
R8SP1 & 8 & 1 & \textbf{0.00622} & 0.9965 \\
R4SP2 & 4 & 2 & 0.00704         & 0.9959 \\
R2SP4 & 2 & 4 & 0.01797         & 0.9892 \\
\bottomrule
\end{tabular}
\end{table}

The configuration in which spatial breadth would dominate is narrow.
Retrieval is independent per token and per head, so different tokens or heads needing spatially distant information is already handled.
The remaining failure mode requires a single token, on a single head, to need fine-grained access to specific keys at positions separated by whole documents that fall in different spatial blocks of the same semantic cluster, and even then only when simple lookup does not suffice (any copy of the relevant key being adequate), \ie when fine-grained non-redundant aggregation across those blocks is required.
We found semantic breadth to dominate on our data; on a domain where such dispersed aggregations proved important, shifting budget from $k_1$ to $k_2$ would be the natural adjustment, which our implementation supports.

\subsection{Cluster Count with Fixed Retrieval}
\label{app:cluster_retrieval_1b}

Table~\ref{tab:1b_cluster_fixed_r} shows the effect of varying cluster count with retrieval count held fixed at R=8.
Higher cluster counts improve accuracy (0.00838 $\to$ 0.00456, a $1.8\times$ error reduction) while reducing retrieval work (8192 $\to$ 2048 keys per query).
This confirms the 96M result (Table~\ref{tab:micro_cluster_retrieval}): finer clustering improves monopole quality enough to more than compensate for retrieving from smaller clusters.

\begin{table}[h]
\centering
\small
\caption{Effect of cluster count with fixed retrieval R=8 (SP1B8k, 1B model, 80 heads, 2 sequences, 64k context).}
\label{tab:1b_cluster_fixed_r}
\begin{tabular}{lcccc}
\toprule
Config & RSE & Correlation & Keys/query \\
\midrule
QK64R8 & 0.00838 & 0.9949 & 8192 \\
QK128R8 & 0.00622 & 0.9965 & 4096 \\
QK256R8 & 0.00456 & 0.9975 & 2048 \\
\bottomrule
\end{tabular}
\end{table}

\subsection{Monopole Runtime Scaling}
\label{app:monopole_scaling}

We measure monopole-only runtime (no retrieval) to characterize the scaling of the novel MuSe components: clustering, summary computation, and causal accumulation.
All timings are for 40 heads on a single node (5 heads per A100).

\paragraph{Sequence Length at Constant Total Tokens}
Table~\ref{tab:monopole_seqlen} shows monopole runtime at constant total tokens (128k) but varying sequence length and batch size.
Longer sequences are modestly slower despite equal total work, likely due to reduced parallelism across sequences or less efficient clustering on larger point sets rather than algorithmic overhead.

\begin{table}[h]
\centering
\small
\caption{Monopole-only runtime at constant total tokens (128k), varying sequence length (QK128B8k, 40 heads).}
\label{tab:monopole_seqlen}
\begin{tabular}{lccc}
\toprule
Config & Sequences & Blocks/seq & Time (ms) \\
\midrule
N=32k  & 4 & 4  & 33.3 \\
N=64k  & 2 & 8  & 36.0 \\
N=128k & 1 & 16 & 41.0 \\
\bottomrule
\end{tabular}
\end{table}

Per-token cost increases modestly with sequence length: 0.26~$\mu$s/token at 32k context to 0.32~$\mu$s/token at 128k context (+23\%).
This is slightly super-linear in sequence length at constant batch size, but the overhead is mild.

\paragraph{Cluster Count Scaling}
Table~\ref{tab:monopole_clusters} shows monopole runtime as cluster count varies at fixed context length.
Increasing clusters from 64 to 256 (4$\times$) increases runtime by only 31\%, indicating that cluster-dependent costs (summaries, accumulation) are not dominant.

\begin{table}[h]
\centering
\small
\caption{Monopole-only runtime varying cluster count (B=16k, N=64k, 40 heads, constant total tokens via batching).}
\label{tab:monopole_clusters}
\begin{tabular}{lcc}
\toprule
Clusters & Time (ms) & vs QK64 \\
\midrule
QK64  & 52.1 & --- \\
QK128 & 55.3 & +6\% \\
QK256 & 68.3 & +31\% \\
\bottomrule
\end{tabular}
\end{table}

If accumulation (which scales as $C^2$) dominated, we would expect 4$\times$ clusters to yield 16$\times$ runtime; the observed 31\% increase confirms that accumulation is a small fraction of total monopole cost.

\paragraph{Sequence Length with Larger Blocks}
Table~\ref{tab:monopole_seqlen_b16k} shows a similar experiment with B=16k blocks and QK64.

\begin{table}[h]
\centering
\small
\caption{Monopole-only runtime at constant total tokens (128k), varying sequence length (QK64B16k, 40 heads).}
\label{tab:monopole_seqlen_b16k}
\begin{tabular}{lccc}
\toprule
Config & Sequences & Blocks/seq & Time (ms) \\
\midrule
N=32k  & 4 & 2 & 49.4 \\
N=64k  & 2 & 4 & 52.1 \\
N=128k & 1 & 8 & 57.9 \\
\bottomrule
\end{tabular}
\end{table}

The pattern is consistent: longer sequences are slower at constant total tokens (+17\% from 32k to 128k), but the overhead is modest and likely attributable to reduced parallelism or clustering efficiency rather than algorithmic complexity.

\subsection{Comparison with MoBA}
\label{app:moba_comparison}

The main text compares \ours{} against MoBA at matched far-field sparsity (8k block diagonal, 64$\times$ sparsity), isolating the quality of far-field approximation.
For completeness, we also compare against MoBA with settings closer to typical usage: 512-token block diagonal with top-8 retrieval from 512-token blocks (8$\times$ far-field sparsity).
This configuration allocates more compute budget to far-field retrieval at the cost of local attention quality.

Table~\ref{tab:moba_extended} shows that even under these favorable conditions, \ours{} maintains an advantage at both scales.
Note that this comparison favors MoBA: our 8k block size is chosen to balance accuracy and runtime with our kernels, not to maximize sparsity.
The 512-block MoBA configuration would be substantially slower than exact attention when implemented with our kernels.

\begin{table}[h]
\centering
\small
\caption{Extended MoBA comparison at 185M and 560M scale. ``Sparsity-matched'' uses 8k block diagonal with 64$\times$ far-field sparsity; ``Budget-matched'' uses 512-token blocks with 8$\times$ far-field sparsity. Cross-entropy loss, lower is better.}
\label{tab:moba_extended}
\begin{tabular}{llcc}
\toprule
\multirow{2}{*}{Scale} & \multirow{2}{*}{Method} & \multicolumn{2}{c}{Test Attention} \\
\cmidrule(lr){3-4}
& & CUDNN & Method \\
\midrule
\multirow{3}{*}{185M} & CUDNN (baseline) & 0.9400 & --- \\
& \ours{} & \textbf{0.9384} & 0.9435 \\
& MoBA (sparsity-matched) & 0.9714 & 1.0027 \\
& MoBA (budget-matched) & 0.9460 & 0.9633 \\
\midrule
\multirow{3}{*}{560M} & CUDNN (baseline) & 0.7745 & --- \\
& \ours{} & \textbf{0.7728} & 0.7746 \\
& MoBA (sparsity-matched) & 0.7958 & 0.8209 \\
& MoBA (budget-matched) & 0.7815 & 0.7858 \\
\bottomrule
\end{tabular}
\end{table}

\subsection{Far-Field Scaling with Context Length}
\label{app:farfield_scaling}

Table~\ref{tab:farfield_scaling} shows approximation quality versus context length with scaled hyperparameters: $\text{QK} \propto \sqrt{N}$, $R \propto \sqrt{N}$ (constant retrieval fraction), $B \propto N$ (constant 8 spatial blocks).
This maintains 64$\times$ far-field sparsity and 1/8 block diagonal fraction across all context lengths.

\begin{table}[h]
\centering
\small
\caption{Far-field approximation quality vs.\ context length with scaled hyperparameters (8 spatial blocks throughout). Relative squared error decreases as context grows. $^*$128k context created by concatenating two 64k sequences.}
\label{tab:farfield_scaling}
\begin{tabular}{rrrr|cc}
\toprule
$N$ & QK & $R$ & $B$ & RSE & Cosine \\
\midrule
\multicolumn{6}{l}{\textit{Series 1:}} \\
4k & 32 & 2 & 512 & 0.00884 & 0.9947 \\
16k & 64 & 4 & 2k & 0.00662 & 0.9958 \\
64k & 128 & 8 & 8k & 0.00622 & 0.9965 \\
\midrule
\multicolumn{6}{l}{\textit{Series 2:}} \\
2k & 16 & 1 & 256 & 0.01701 & 0.9899 \\
8k & 32 & 2 & 1k & 0.01096 & 0.9929 \\
32k & 64 & 4 & 4k & 0.00863 & 0.9946 \\
128k$^*$ & 128 & 8 & 16k & 0.00723 & 0.9956 \\
\bottomrule
\end{tabular}
\end{table}

Approximation error decreases substantially with scale (0.017 $\to$ 0.007 from 2k to 128k), implying that the compute required to achieve a given error level grows sub-quadratically in the far-field.
We do not determine the precise exponent, which would likely be data-dependent.
The block diagonal cost remains $O(N^2/S)$; for very long contexts where this dominates, recursive application of \ours{} would reduce it further.

\section{Additional Results}
\label{app:additional}


\subsection{Effective Cluster Count Comparison}
\label{app:effective_clusters}

Tables~\ref{tab:effective_muse} and~\ref{tab:effective_noq} compare MuSe and the no-query-clustering ablation by computing effective cluster counts.
For each configuration, we interpolate (or extrapolate) using a power-law fit to determine what cluster count the other method would require to achieve the same error.
Both methods follow approximate power laws: MuSe with exponent $d \approx -0.89$ and no-query-clustering with $d \approx -0.59$.
The steeper slope for MuSe means its advantage grows with cluster count.

\begin{table}[h]
\centering
\small
\caption{MuSe configurations with equivalent no-query-clustering cluster counts. R = QK/16 throughout. Effective cluster count interpolated via power-law fit. Multiplier shows how many more clusters the no-query-clustering method would need.}
\label{tab:effective_muse}
\begin{tabular}{lcccc}
\toprule
QK & RSE & Corr & Eff.\ No-Q & Mult.\ \\
\midrule
16   & 0.0380 & 0.9775 & 82    & 5.1$\times$ \\
32   & 0.0210 & 0.9878 & 191   & 6.0$\times$ \\
64   & 0.0114 & 0.9935 & 568   & 8.9$\times$ \\
128  & 0.0062 & 0.9965 & 1179  & 9.2$\times$ \\
256  & 0.0033 & 0.9980 & 2518  & 9.8$\times$ \\
512  & 0.0017 & 0.9990 & 5481  & 10.7$\times$ \\
\bottomrule
\end{tabular}
\end{table}

\begin{table}[h]
\centering
\small
\caption{No-query-clustering configurations with equivalent MuSe cluster counts. R = QK/16 throughout. Multiplier shows how many fewer clusters MuSe needs to achieve the same error.}
\label{tab:effective_noq}
\begin{tabular}{lcccc}
\toprule
QK & RSE & Corr & Eff.\ MuSe & Mult.\ \\
\midrule
16   & 0.0991 & 0.9464 & 6   & 2.7$\times$ \\
32   & 0.0665 & 0.9640 & 9   & 3.6$\times$ \\
64   & 0.0428 & 0.9775 & 13  & 4.9$\times$ \\
128  & 0.0284 & 0.9856 & 19  & 6.7$\times$ \\
256  & 0.0191 & 0.9904 & 36  & 7.1$\times$ \\
512  & 0.0129 & 0.9935 & 57  & 9.0$\times$ \\
\bottomrule
\end{tabular}
\end{table}

\subsection{Downstream Evaluation}
\label{app:downstream}

To verify that \ours{} does not degrade downstream task performance, we evaluate 1B models trained on scientific PDFs using the lm-eval harness on ARC-Easy and SciQ.
Table~\ref{tab:downstream} shows that \ours{}-trained models perform comparably to CUDNN-trained models, with all differences within one standard error.
Both models are well above the 25\% random baseline, confirming that the approximation does not impair learned representations.

\begin{table}[h]
\centering
\small
\caption{Downstream evaluation on ARC-Easy and SciQ (1B model trained on scientific PDFs). All differences are within standard error ($\sim$0.01).}
\label{tab:downstream}
\begin{tabular}{llcc}
\toprule
Benchmark & Metric & CUDNN & \ours{} \\
\midrule
ARC-Easy & acc & 0.479 & 0.487 \\
ARC-Easy & acc\_norm & 0.431 & 0.436 \\
SciQ & acc & 0.766 & 0.759 \\
SciQ & acc\_norm & 0.663 & 0.659 \\
\bottomrule
\end{tabular}
\end{table}

\subsection{RULER Results}
\label{app:ruler_full}

Table~\ref{tab:ruler} reports per-task RULER accuracy at 64k context, for both domains and both training methods.
As discussed in Section~\ref{sec:retrieval_eval}, both \ours{}- and CUDNN-trained models score poorly on most variants (the code models near zero on all but single-needle retrieval), which we attribute to domain shift and the limited instruction-following of our small base models rather than to a retrieval deficit.

\begin{table}[h]
\centering
\small
\caption{\textbf{Per-task RULER NIAH accuracy at 64k context.} Accuracy $\pm$ binomial std.\ error, $n=500$ per cell. lm-eval-harness reports no stderr for these tasks; we compute $\sqrt{p(1-p)/n}$. Both methods were run on all tasks for both domains. See Section~\ref{sec:retrieval_eval} for interpretation.}
\label{tab:ruler}
\begin{tabular}{lcc}
\toprule
Task & CUDNN & \ours{} \\
\midrule
\multicolumn{3}{l}{\emph{Scientific PDF domain}} \\
\texttt{niah\_single\_1}   & 61.0 $\pm$ 2.2\%          & \textbf{85.0 $\pm$ 1.6\%} \\
\texttt{niah\_single\_2}   & \textbf{46.6 $\pm$ 2.2\%} & 26.6 $\pm$ 2.0\% \\
\texttt{niah\_multikey\_1} & \textbf{50.4 $\pm$ 2.2\%} & 28.0 $\pm$ 2.0\% \\
\texttt{niah\_multiquery}  & \textbf{23.8 $\pm$ 1.9\%} & 15.4 $\pm$ 1.6\% \\
\texttt{niah\_multivalue}  & \textbf{20.9 $\pm$ 1.8\%} & 15.9 $\pm$ 1.6\% \\
\midrule
\multicolumn{3}{l}{\emph{Code domain}} \\
\texttt{niah\_single\_1}   & 62.8 $\pm$ 2.2\%          & \textbf{70.6 $\pm$ 2.0\%} \\
\texttt{niah\_single\_2}   & \phantom{0}0.0\%          & \phantom{0}1.4 $\pm$ 0.5\% \\
\texttt{niah\_multikey\_1} & \phantom{0}0.0\%          & \phantom{0}2.2 $\pm$ 0.7\% \\
\texttt{niah\_multiquery}  & \phantom{0}0.0\%          & \phantom{0}0.0\% \\
\texttt{niah\_multivalue}  & \phantom{0}4.6 $\pm$ 0.9\% & \phantom{0}0.0\% \\
\bottomrule
\end{tabular}
\end{table}

\subsection{Custom NIAH Accuracy by Needle Depth}
\label{app:niah_depth}

Table~\ref{tab:niah_depth} resolves the custom-NIAH uid-level exact match of Section~\ref{sec:retrieval_eval} by needle depth, for the fixed-depth variants (single-needle and multi-key; the multi-query variants place needles at randomly sampled depths and so are not bucketed here).
\ours{}-trained accuracy remains above 89\% at every depth.
The exact-attention baseline, by contrast, degrades sharply toward the middle depths on the harder multi-key $4\times$ variant (63.6\% at depth 0.25 versus 96.7\% at 0.75).

\begin{table}[h]
\centering
\footnotesize
\setlength{\tabcolsep}{3pt}
\caption{Custom code NIAH uid-level exact match (\%) by needle-depth bucket, for the fixed-depth variants. CUDNN denotes exact-attention training and \ours{} denotes \ours{} training; all evaluation uses exact attention.}
\label{tab:niah_depth}
\begin{tabular}{lcccccc}
\toprule
& \multicolumn{2}{c}{Single-needle} & \multicolumn{2}{c}{Multi-key $2\times$} & \multicolumn{2}{c}{Multi-key $4\times$} \\
\cmidrule(lr){2-3}\cmidrule(lr){4-5}\cmidrule(lr){6-7}
Depth & CUDNN & \ours{} & CUDNN & \ours{} & CUDNN & \ours{} \\
\midrule
0.05 & 96.8 & 89.1  & 92.0 & 89.8  & 85.1 & 90.7 \\
0.15 & 97.5 & 98.6  & 91.6 & 99.6  & 77.3 & 98.1 \\
0.25 & 98.2 & 99.7  & 86.2 & 100.0 & 63.6 & 98.9 \\
0.35 & 97.2 & 100.0 & 84.4 & 100.0 & 67.3 & 99.3 \\
0.45 & 97.9 & 100.0 & 90.2 & 99.6  & 75.5 & 98.9 \\
0.55 & 97.5 & 99.3  & 93.1 & 98.9  & 83.3 & 97.0 \\
0.65 & 97.5 & 99.3  & 98.6 & 98.2  & 90.7 & 94.8 \\
0.75 & 99.3 & 99.7  & 99.3 & 100.0 & 96.7 & 98.5 \\
0.85 & 98.2 & 99.7  & 99.3 & 100.0 & 94.1 & 100.0 \\
\bottomrule
\end{tabular}
\end{table}

\subsection{Fine-tuning to Remove Adaptation}
\label{app:finetuning}

At 1B scale on the code domain, the \ours{}-trained model shows minor adaptation to the approximation: when evaluated with CUDNN attention, loss is 0.7108 compared to the baseline of 0.7026.
We investigate whether brief fine-tuning with exact attention removes this adaptation.

Table~\ref{tab:finetuning} shows the progression of MuSe$\to$CUDNN evaluation loss during CUDNN fine-tuning.
The adaptation gap closes rapidly: after just 26M tokens (0.1\% of pretraining), the model not only matches but \emph{beats} the CUDNN baseline (0.6994 vs 0.7026).
Continued fine-tuning yields diminishing returns, with 93\% of the improvement occurring in the first 0.1\% of fine-tuning tokens.
This confirms that adaptation effects are shallow and easily removed, making the recommended workflow straightforward: pretrain with \ours{} for speedup, then briefly fine-tune with exact attention before deployment.
As a preliminary alternative that avoids the fine-tuning step entirely, the ``Aggregation Weight Gradient Flow'' paragraph of Appendix~\ref{app:clustering} reports a training-time gradient modification that removes the adaptation gap directly.

\begin{table}[h]
\centering
\small
\caption{Fine-tuning progression on 1B code model. MuSe$\to$CUDNN evaluation loss during CUDNN fine-tuning.}
\label{tab:finetuning}
\begin{tabular}{rrcl}
\toprule
MTok & \% Pretrain & Loss & Note \\
\midrule
0 & 0\% & 0.7108 & Before fine-tuning \\
26 & 0.1\% & 0.6994 & Beats baseline (0.7026) \\
210 & 0.9\% & 0.6985 & Continued improvement \\
\bottomrule
\end{tabular}
\end{table}

\section{Distributed Training Considerations}
\label{app:distributed}

\ours{} is compatible with standard distributed training strategies.

\paragraph{Sequence parallelism}
In Megatron-style sequence parallelism, attention heads are sharded across devices, with the full sequence gathered for each head before attention.
From the perspective of \ours{}, each device simply approximates fewer heads over the complete sequence---the distributed origin of sequence fragments is invisible to the approximation.
Our pretraining experiments use this approach.

\paragraph{Ring attention}
Ring attention distributes the sequence across devices, rotating key/value blocks around a ring.
The block-diagonal (local, causal) attention is naturally handled by \ours{}'s spatial blocking.
For off-diagonal (remote, non-causal) blocks, \ours{} extends naturally: either retain spatial blocking with the causal mask removed, or treat remote context as a single spatial block with purely semantic sparsity.
We leave detailed investigation of these extensions to future work.

\section{Code Availability}
\label{app:code}

The research code used for the experiments in this paper is available as the supplementary material accompanying the submission on OpenReview.
It is provided to document the exact implementation used for the experiments reported here, and for reproducibility, not as a maintained or general-purpose implementation of the method.

\end{document}